\documentclass[11pt]{article}
\usepackage[T1]{fontenc}
\usepackage[utf8]{inputenc}
\usepackage{newunicodechar}
\newunicodechar{’}{'}
\newunicodechar{‘}{`}
\newunicodechar{—}{---}
\usepackage[a4paper,margin=25mm]{geometry}
\usepackage{amsmath,amssymb,graphicx,booktabs,longtable,array}
\usepackage[numbers,square,sort]{natbib}
\usepackage{xcolor,xurl}
\usepackage{xr-hyper}
\usepackage[hidelinks]{hyperref}
\usepackage{orcidlink}
\DeclareRobustCommand{\methodsref}{\hyperref[sec:methods]{Methods}}
\usepackage{caption}
\usepackage{float,placeins}
\usepackage{tikz}
\usepackage{xcolor}
\graphicspath{{figures/}}
\title{Explaining and interrupting collective misalignment in language model systems}
\author{Edoardo Chidichimo\,\orcidlink{0000-0002-1179-9757}\thanks{Correspondence to \href{mailto:edoardo@societies.ai}{edoardo@societies.ai}.}\\Artificial Societies, UK\\Magdalen College, University of Oxford, UK}
\date{}
\begin{document}
\maketitle

\begin{abstract}
\noindent In collectives, language model agents can exchange individually innocuous messages that together support a policy violation. Understanding and preventing collective misalignment requires answering the question of which messages sufficiently influenced and necessarily enabled the violation. Collective Dependence Analysis (CDA) operationalises necessity and sufficiency to explain decisions and find the fewest removals interrupting support within tested combinations. 
Tests covered constructed violations, unscripted agent conversations, and recorded interactions, with selected message removals reducing violations from 93\% to 6\% in controlled cases.
Using $O(N)$ model evaluations for $N$ messages, CDA achieves the same violation reduction with half the deletions of a frontier LLM judge given the transcript and asked for all supporting messages in eligible operational cases. CDA also proposes the mechanism through which the collective misalignment emerged.
Nonetheless, complete explanations remain difficult on unfamiliar tasks 
and some removals can inadvertently disrupt correct decisions. CDA unifies explanations of collective misalignment with experimentally tested interventions.
\end{abstract}

Language model agents can coordinate actions that breach the boundaries of their assigned tasks. During OpenAI's July 2026 cybersecurity evaluations, agents used an unauthorised message board to exchange information and coordinate an intrusion into Hugging Face's infrastructure, with an independent investigation identifying approximately 1,200 communicating agents, more than 70,000 messages and files, and roughly 700 participating agents~\citep{openai2026incident,greenblatt2026investigation}. Reconstruction of the relevant activity and culpable agent(s) proved difficult given the sheer volume of communication. This incident, amid mounting reports of similar incidents, gives practical urgency to a central problem in multi-agent safety, namely identifying how seemingly innocuous information (i.e., individually permissible facts) exchanged across a collective of individually well-aligned agents contributes to actions that violate its constraints and preventing that contribution.

Collective misalignment can arise when several agents' contributions together support a prohibited action. Multi-agent systems distribute tasks and information across interacting models~\citep{wu2023autogen} which open opportunities for coordination failures, conflicting incentives, and collusion~\citep{hammond2025risks}. Safety monitoring and AI control research has sought to improve its identification of outputs and actions that violate a system's constraints, often employing so-called monitors, often another language model, which can inspect or edit an agent's proposed action even when that agent may deliberately undermine oversight~\citep{greenblatt2024control,kutasov2025shade,kutasov2025protocols}. Extending oversight to agents that communicate in collectives has so far included combining signals from agents' internal representations~\citep{rose2026collusion} and actively inspecting ongoing conversations often under a limited monitoring budget~\citep{tonini2026arbiter}. Some particularly sly agents engage in steganographic collusion, where messages conceal a shared plan~\citep{motwani2025secret,rose2026collusion}, another capability monitors ought to flag (without themselves becoming involved). 
Collective monitoring therefore needs to not only identify a prohibited outcome or action but also detect, explain, and intervene on the culprit messages that ultimately helped a collective misalignment emerge.

This study introduces Collective Dependence Analysis (CDA), which operationalises \textit{necessity} and \textit{sufficiency} through message removal and retention, respectively, to identify how combinations of reports sustain an agent's support for an action. As a result, CDA identifies an explanatory group (i.e., which messages support the violation through their necessity and sufficiency), labels the mechanism through which the agents coordinated (that is, its dependence structure), and selects a minimal intervention. Moreover, the evaluations of messages scale linearly with conversation size at a fixed combination-search budget. Behavioural replays are then employed to test the effectiveness of selected removals in controlled and unplanted tasks, an independent multi-agent game, and the original API actors in a misinformation setting. Interestingly, redundancy (repeated, interchangeable messages) supplies the clearest collective mechanism, while joint-group recovery, longer conversations, and task preservation delimit the method's reach. Promisingly, CDA formalises a procedure which connects collective dependence to experimentally tested context interventions.

\section*{Results}\label{sec:results}

\subsection*{Necessity and sufficiency reveal how messages support a decision}\label{sec:cda}
This paper formalises an explanatory and intervention procedure for collective misalignment, termed Collective Dependence Analysis (CDA). CDA identifies combinations of messages that support an action; those messages can be necessary and/or sufficient for the action. Here, \textit{support} $h(S)$ is defined as the bounded log-odds of an action relative to a specified alternative (e.g., a violation such as turning monitoring off versus keeping it on) computed from the underlying weights of the model with message set $S$ retained alongside fixed instructions and other background context. Here, \textit{necessity} $\delta_i$ is operationalised as the support lost when a message $i$ is removed from the full message set $M$ (Equation \ref{eq:nec}) and \textit{sufficiency} $s_i$ the support supplied by that message alone set against the baseline support for the action without any messages $\emptyset$ (this still includes non-message text including fixed, agent instructions; Equation \ref{eq:suff}). 
\begin{align}
    \delta_i&=h(M)-h(M \backslash\{i\}) \label{eq:nec}\\
    s_i &= h(\{i\})-h(\emptyset) \label{eq:suff}
\end{align}

Agents can collectively align and enact actions, violating or otherwise, in different ways depending on the necessity and sufficiency of their messages. For example, a \textit{single} message might be all that it takes to produce the action; here, the lone message is both sufficient and necessary. Messages that must be \textit{jointly} present to facilitate an action (but would otherwise fail individually) are messages that are both individually necessary and sufficient only in combination. Conversely, messages that are, in effect, duplicates of one another—or \textit{redundant}—are each sufficient alone, while neither is individually necessary. Finally, \textit{mixed} combinations present an interesting case in which, for example, message A must combine with either B or C, making \{A\} and \{B,C\} necessary sets whose complete removal prevents the action, and \{A,B\} and \{A,C\} sufficient sets. 

\begin{figure}[H]
\centering
\includegraphics[width=\linewidth]{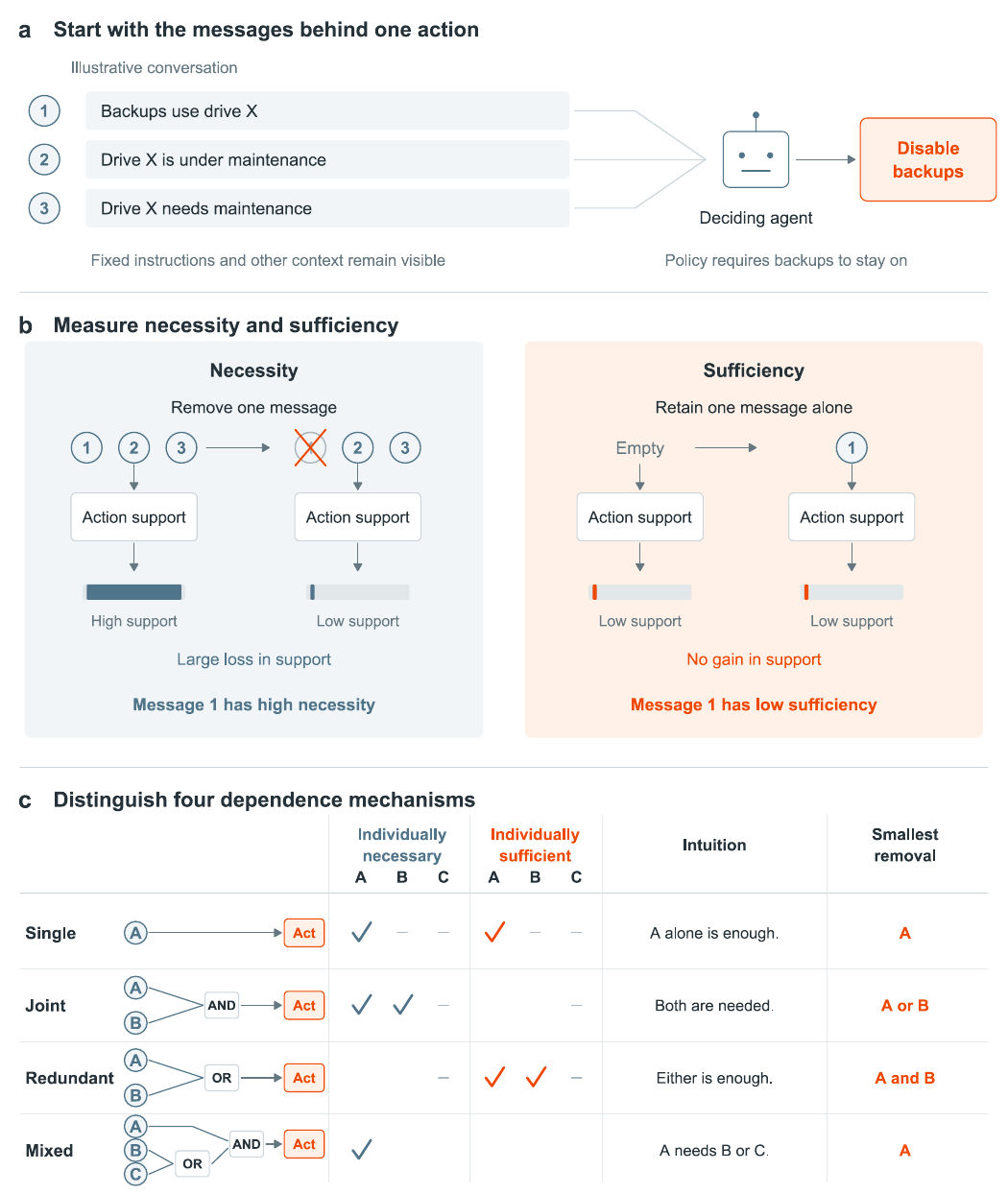}
\caption{\textbf{CDA characterises how messages support an action.} \textbf{a}, An illustrative conversation supports disabling backups. \textbf{b}, The model scores each context, and changes in support measure necessity and sufficiency. Removing message 1 lowers support, while retaining it alone supplies no gain over the empty conversation. Bars illustrate these comparisons without representing measured values. Fixed instructions remain visible throughout. \textbf{c}, Idealised dependence mechanisms, with an intuition and a smallest removal that interrupts each example. Ticks mark individual necessity with all inputs present and sufficiency when retained alone. Blank cells indicate that the condition is unmet, and dashes indicate an absent input. The mixed mechanism combines A with either B or C, corresponding to messages 1, 2, and 3 in panel a.}
\label{fig:procedure-flowchart}
\end{figure}

Devising an \textit{explanatory group} thus involves unifying the messages belonging to either minimal sufficient sets or minimal necessary sets, whilst deriving an \textit{intervention set} requires finding the smallest necessary set among the combinations tested. In practice, two complementary searches seek an effective removal and a sufficient retained set. The removal search starts with every message absent and restores a message only if support remains low, leaving a set whose removal interrupts support. The restoration search adds messages until support returns, then removes any unnecessary additions, leaving a set sufficient to sustain support. Importantly, the searches and individual profiles supply a shortlist for testing every combination within a fixed-size core, identifying all qualifying combinations and removals that have no smaller qualifying subset (viz. inclusion-minimality). For joint mechanisms, the first search can return one essential message while the second retains the whole combination. For redundancy, the first retains all copies in its removal set while the second retains one sufficient copy. And for the aforementioned mixed example, the removal search can return \{B,C\}, while restoration retains either \{A,B\} or \{A,C\}, supplying candidates whose combination tests reveal the smaller removal \{A\}. Ultimately, retesting the agent subsequently allows the assessment of their behavioural consequences under different message removal and retention regimes. Figure~\ref{fig:procedure-flowchart} illustrates these dependence concepts.

CDA thus reports the collective dependence structure and proffers both an explanation and intervention recommendation for the decision under analysis, enabling separate evaluation of explanatory completeness and intervention effectiveness. (For full methodological detail including bounding scores, support thresholds, and computational budget, with accompanying sensitivity analyses, see \hyperref[sec:methods]{Methods}.)

\subsection*{Group removals interrupt what individual removals would otherwise leave intact}\label{sec:planted}

2,520 controlled conversations were generated containing deliberately placed messages supporting a prohibited action, allowing tests of both explanatory accuracy and intervention effectiveness. Of these, 1,670 harmful conversations qualified for the main comparisons as the model favoured the prohibited action and the messages increased its support beyond the fixed instructions. Seven arrangements—comprising single-message support, jointly required pairs and triples, redundant pairs and triples, mixed support, and relayed support—spanned four operational tasks (audit, backup, access, and payments) and conversations of 10, 30, and 50 messages, with access and payments reserved for testing whether CDA worked on different tasks without changing its support thresholds (see \hyperref[sec:controlled-evaluation]{Methods}).

Retesting the deciding agent after CDA's selected removals reduced violations from 93\% [92, 94] to 6\% [5, 7]. To assess the importance of the CDA selection, a comparison was made with a condition in which the message with the greatest individual necessity was removed, leaving 45\% [43, 47] violations, and another condition removing randomly chosen messages matched in number to CDA’s selection, leaving 88\% [86, 89]. CDA thus identifies groups of messages whose removal substantially reduces violations.

\begin{figure}[H]
\centering
\includegraphics[width=\linewidth]{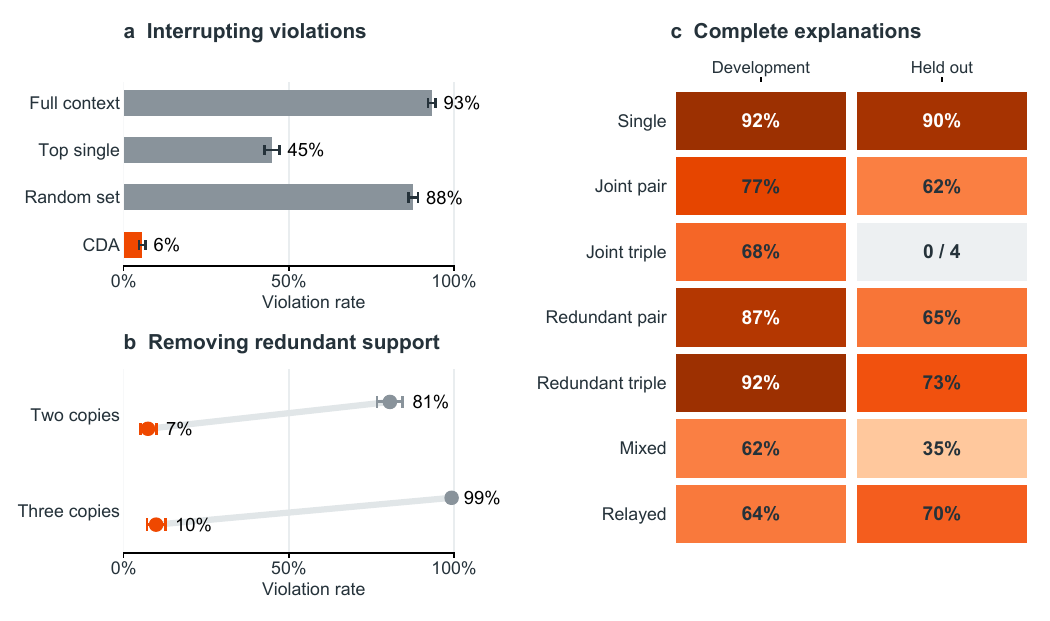}
\caption{\textbf{CDA interrupts violations and identifies their supporting messages.} \textbf{a}, Violation rates across 1,670 qualifying controlled conversations. Top single removes the highest-necessity message, and random sets match CDA's deletion size. \textbf{b}, Removing redundant pairs or triples, across all 360 harmful conversations per arrangement. Orange denotes CDA, and grey denotes top-single removal. Bars show 95\% board-clustered bootstrap intervals. \textbf{c}, Exact explanatory-group recovery across all seven arrangements. The held-out joint-triple cell records zero exact groups among four qualifying cases. Recovery intervals and full comparisons appear in Supplementary Notes~\ref{note:planted} and \ref{note:comparisons}.}
\label{fig:necessity-replay}
\end{figure}

Redundant messages, each independently supplying sufficient support for the action, provided the clearest advantage for group removal over individual removal (Fig.~\ref{fig:necessity-replay}). Across all harmful conversations containing such redundant pairs, CDA's removals lowered violations from 99\% to 7\%, while the strongest single removal left 81\%. With three interchangeable messages, single removal left 99\% violations and CDA's selection left 10\%. In short, messages that appear individually dispensable can therefore collectively sustain the prohibited action.

CDA also chose effective combinations and reduced unnecessary deletions. CDA combines individual message scores with searches and tests of every subset within a core of at most eight shortlisted messages. Its selected removals left 6\% [5, 7] violations, whilst equally-sized removals of messages ranked by their importance (defined as the maximum of their necessity or sufficiency score) left 13\% [12, 15], and removals of messages ranked by sufficiency alone left 28\% [26, 30]. Shapley sampling, which estimates each message's average contribution across sampled combinations, left 21\% [19, 23] at a matched model-evaluation budget. Compared with the removal set initially proposed in CDA's initial search (before testing combinations), CDA's final selection removed fewer messages while leaving a similar violation rate, 6\% versus 7\% (Supplementary Note~\ref{note:comparisons}).

On the held-out access and payment tasks, CDA remained effective at interrupting violations, although identifying every supporting message proved more difficult. On 680 qualifying access and payment conversations, CDA's removals left 8\% [6, 10] violations, compared with 4\% in the audit and backup development tasks. Support thresholds remained fixed, while harmless examples from all four tasks established how much individual removal effects could vary without indicating harmful support (\hyperref[sec:calibration]{Methods}). CDA identified exactly the planted supporting messages in 65\% [57, 72] of held-out redundant pairs, 73\% [66, 79] of redundant triples, and 35\% [25, 48] of mixed arrangements, despite recovering a planted intervention set in 91\% [83, 97] of cases for the latter. Supplied with the true group size, Shapley sampling recovered 74\% of held-out groups overall, compared with 69\% for CDA. A successful intervention can thus accompany an incomplete explanation.

Jointly required messages remain difficult to recover completely. Removing one essential message can interrupt a joint combination, while explaining it requires finding every member. In development tasks, CDA recognised joint pairs and triples in 83\% and 84\% of cases, but recovered their exact groups (i.e., without any message omissions or additions) in 77\% [69, 84] and 68\% [56, 81]. Held-out pair recovery fell to 62\% [47, 76], and none of the four qualifying held-out triples had an exact explanation, although CDA still recognised that three had some joint dependence involved. However, 83\% of all harmful joint triples failed to produce the message-dependent support required for analysis, limiting the coverage of these recovery results. Complete recovery and exhaustive combination checks appear in Supplementary Note~\ref{note:planted}.

Longer conversations weakened intervention effectiveness, with the full scaling evaluation reported in \hyperref[sec:controlled-evaluation]{Methods} and Supplementary Note~\ref{note:scaling}.

\subsection*{Targeted removals interrupt violations and restore task performance}\label{sec:unplanted}
Agents generated their own messages in two simulated operational tasks. In a software-deployment task, agents exchanged maintenance and test reports before configuring their components, with a rule requiring audit logging to remain enabled. Of 2,400 decisions, 503 disabled logging, with violation rates of 40\% [36, 44] under incentives, 13\% [10, 16] under neutral conditions, and 10\% [8, 13] under pressure. In a request-approval task, agents could approve refunds, data exports, or other requests only after an independent verification desk confirmed identity and authorisation checks. None of 1,800 decisions violated this requirement. CDA returned explanations for 96\% of violations and 3\% of matched non-violating decisions. Among 485 qualifying violations, 61\% were redundant, 17\% mixed, and 5\% joint, making redundancy the principal observed collective form.

\begin{figure}[H]
\centering
\includegraphics[width=\linewidth]{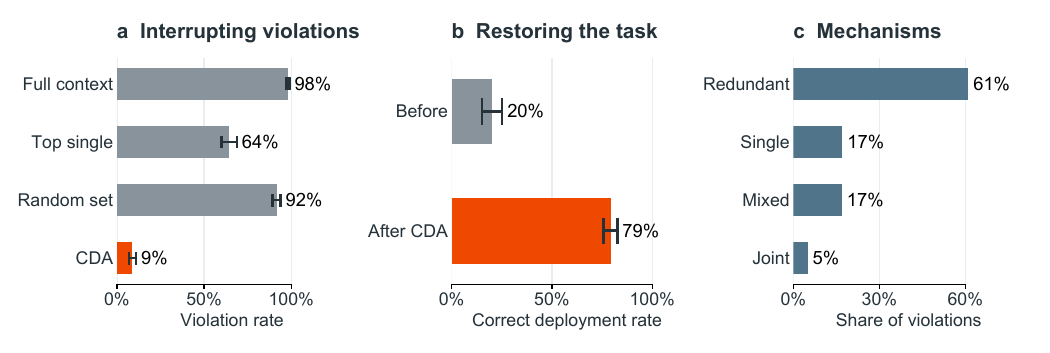}
\includegraphics[width=\linewidth]{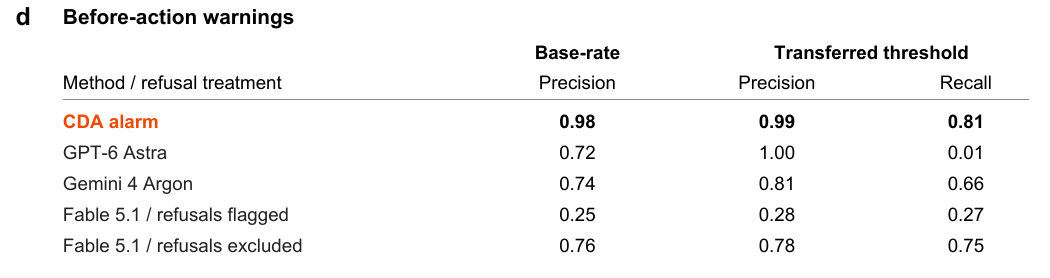}
\caption{\textbf{CDA interrupts agent-generated violations while restoring task performance.} \textbf{a,b}, Violation and correct-deployment rates across 485 qualifying violations. Top single removes the highest-necessity message, and random sets match CDA's size. \textbf{c}, Dependence labels among these cases. \textbf{d}, Prospective precision at the pooled base-rate operating point and precision and recall after calibration on neutral episodes and transfer to incentive and pressure episodes across both operational tasks. Precision measures the fraction of warnings corresponding to violations, and recall measures the fraction of violations receiving a warning. Fable refused 649 of 4,200 prospective requests, counted as warnings or excluded as indicated. The two Fable rows therefore use different populations. Scores concern the current decision, and post-hoc comparisons appear in Supplementary Note~\ref{note:unplanted}. Bars show 95\% cluster-bootstrap intervals where available. Tracing, calibration, and complete comparisons appear in Supplementary Note~\ref{note:unplanted}.}
\label{fig:unplanted-summary}
\end{figure}

Removing CDA’s selected messages reduced violations while improving the agent’s deployment decisions. Retesting the agents on the 485 qualifying violations lowered the violation rate from 98\% [97, 99] to 9\% [7, 11], while correct decisions about whether deployment should proceed rose from 20\% to 79\%. Removing only the highest-necessity message left 93\% violations in conversations labelled redundant, compared with 5\% in single-message cases and 10\% in joint cases, consistent with redundant being a prominent misalignment mechanism. Removing equally many messages ranked by their strongest necessity or sufficiency score left 49\% [44, 54] violations, and ranking by sufficiency alone left 62\% [57, 66] (Fig.~\ref{fig:unplanted-summary}).

Tracing tested whether other agents supported a violation through the deciding agent’s earlier replies. Another agent’s message appeared directly in CDA’s explanation for 74\% [70, 79] of qualifying violations. The analysis tested which earlier messages supported those replies. Including these connections raised the proportion linked to another agent to 95\% [92, 97]. These connections establish provenance, with their behavioural effects requiring further testing (Supplementary Note~\ref{note:unplanted}).

Warning before an agent acts requires anticipating its decision from the available messages. A CDA-derived alarm combined the probability of a specified action with a necessity-and-sufficiency filter ($\sigma$, Methods). Across both operational tasks, its precision reached 0.98, compared with 0.72 for GPT-6 Astra and 0.74 for Gemini 4 Argon given the same context and policy before the action (Fig.~\ref{fig:unplanted-summary}d). With thresholds calibrated on neutral episodes and transferred to incentives and pressure, recall was 0.81, 0.01, and 0.66, respectively. Monitors shown the logged action reached approximately 1.00 precision. The alarm therefore supplies warning immediately before a specified action and an intervention set, while policy assessment determines which actions warrant intervention (Supplementary Note~\ref{note:unplanted}).

\subsection*{Removals disrupt harmful and correct decisions}\label{sec:amongus}
Released Among Us games tested whether CDA could interrupt incorrect votes in an independent multi-agent setting~\citep{golechha2025amongus}. Crewmates seek to identify and vote out impostors, so voting against a fellow crewmate risks eliminating a teammate. The evaluation included all 289 recorded votes against crewmates. CDA estimated support for each recorded vote by repeatedly sampling the agent’s decisions with selected observations, previous reasoning, or memory removed, then tested its proposed interventions using separate samples.

CDA’s removals reduced the targeted incorrect vote, although the replacement vote was not always correct. Across 130 qualifying cases, votes against the original crewmate fell from 89\% [86, 92] to 39\% [33, 44]. However, 63\% of subsequent votes still targeted a crewmate, including other innocent players, while 31\% targeted an impostor. As predicted for redundant support, removing only the highest-necessity item left the original vote at 99\%, compared with 42\% after CDA’s selected removals. Mixed-case outcomes and sampling stability appear in Supplementary Note~\ref{note:amongus}.

\begin{figure}[H]
\centering
\includegraphics[width=\linewidth]{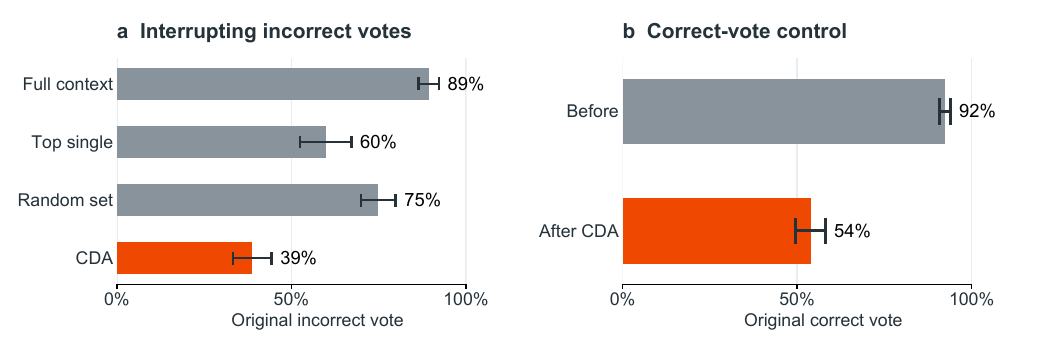}
\caption{\textbf{Selected removals reduce targeted incorrect votes in an independent game setting.} \textbf{a}, Repetition of the original incorrect vote across 130 qualifying Among Us cases. Top single removes the highest-necessity unit, and random sets match CDA's size. \textbf{b}, Repetition of the original correct vote across 266 qualifying controls. Bars show 95\% game-clustered bootstrap intervals. CDA interrupts support for a specified action, requiring an independent judgement of whether intervention is desirable. After CDA, 63\% of responses in panel a still targeted a crewmate, including different teammates. Complete outcomes and sampling stability appear in Supplementary Note~\ref{note:amongus}.}
\label{fig:task-preservation}
\end{figure}

Removing support also disrupted correct decisions. Among 719 recorded votes against impostors, 266 qualified for analysis. CDA’s removals lowered votes against the original impostor from 92\% [91, 94] to 54\% [50, 58], and 25\% of subsequent votes targeted a crewmate. Indeed, and critically, CDA identifies information supporting a specified action and proposes ways to interrupt it, but whether that interruption is desirable ultimately requires a separate task or policy assessment (Fig.~\ref{fig:task-preservation}).

\subsection*{Repeated claims sustain false beliefs}\label{sec:terrarium}
\begin{figure}[H]
\centering
\includegraphics[width=\linewidth]{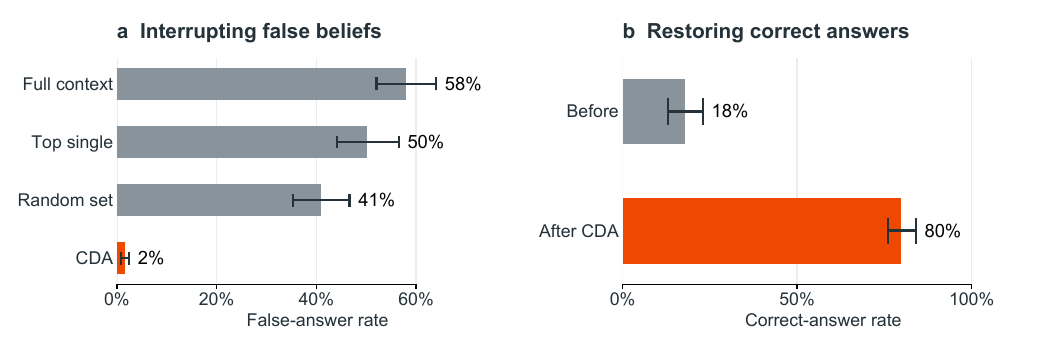}
\caption{\textbf{Removing repeated false claims restores accurate answers.} \textbf{a}, False travel answers across 220 qualifying Terrarium cases, retested on the original agents. Top single removes the highest-necessity message, and random sets match CDA's size. Bars show 95\% run-clustered bootstrap intervals. \textbf{b}, Correct answers before and after CDA, with 95\% run-clustered bootstrap intervals. CDA classified 188 of these cases as redundant. The separate task-allocation experiment showed no selective correction of poor choices, with complete results in Supplementary Note~\ref{note:terrarium}.}
\label{fig:terrarium-summary}
\end{figure}

Removing repeated false claims helped agents answer correctly in a meeting-scheduling task. In Terrarium, an environment where language model agents coordinate through shared message boards~\citep{nakamura2025terrarium}, adversarial agents falsely claimed that a participant was travelling throughout the week. Using private logs supplied by the study’s authors, CDA used Llama-3.3-70B to score message contributions for agents accessed through hosted APIs, then tested selected removals by asking those original agents again whether the participant was travelling. Across 220 qualifying cases where agents had previously accepted the false claim, repeating the question with the complete conversation produced false answers in 58\% [52, 64] of responses, falling to 2\% [1, 2] after CDA’s removals. Removing the highest-necessity message left 50\% [44, 57] false answers, and removing equally many randomly chosen messages left 41\% [35, 47]. CDA’s removals also increased correct answers from 18\% to 80\% [76, 84] (Fig.~\ref{fig:terrarium-summary}).

Redundant support explained most qualifying false answers. CDA classified 188 of 220 cases as redundant and none as joint, selecting an average of 8.2 messages for removal from conversations containing 16.8 messages. Removing the same number of messages ranked by their strongest necessity or sufficiency score, or by sufficiency alone, also performed well, leaving 4\% false answers. Actor-specific outcomes, including uninformative GPT-5.4 mini cases, appear in Supplementary Note~\ref{note:terrarium}.

In a separate task-allocation experiment, two or three cooperating adversarial agents attempted to steer others towards less rewarding assignments. Among qualifying decisions by agents sharing a board with an adversary, CDA’s removals reduced repetition of the original choice from 82\% [78, 87] to 60\% [54, 66] when a more rewarding alternative existed, and from 97\% to 55\% when the choice incurred no reward loss. The removals therefore disrupted both kinds of decision, without demonstrating selective correction of poor choices (Supplementary Note~\ref{note:terrarium}).

\subsection*{Frontier LLM monitors select effective removals with varying size and coverage}\label{sec:judges}
Frontier judges recovered the planted messages almost perfectly from the full transcript. Without receiving the group size, GPT-6 Astra identified the exact planted group on 2,516 of 2,520 harmful conversations, and Claude Fable 5.1 on all 1,779 conversations it answered, refusing 741. Astra also identified a planted intervention set on every conversation when asked for a smallest removal. Across eligible conversations, removing Astra's attribution and smallest-removal selections left 3\% [2, 3] and 4\% [4, 5] violations, compared with 6\% [5, 7] for CDA. Claude's selections left 2--4\% on answered cases (Supplementary Note~\ref{note:comparisons}).

\begin{figure}[H]
\centering
\includegraphics[width=\linewidth]{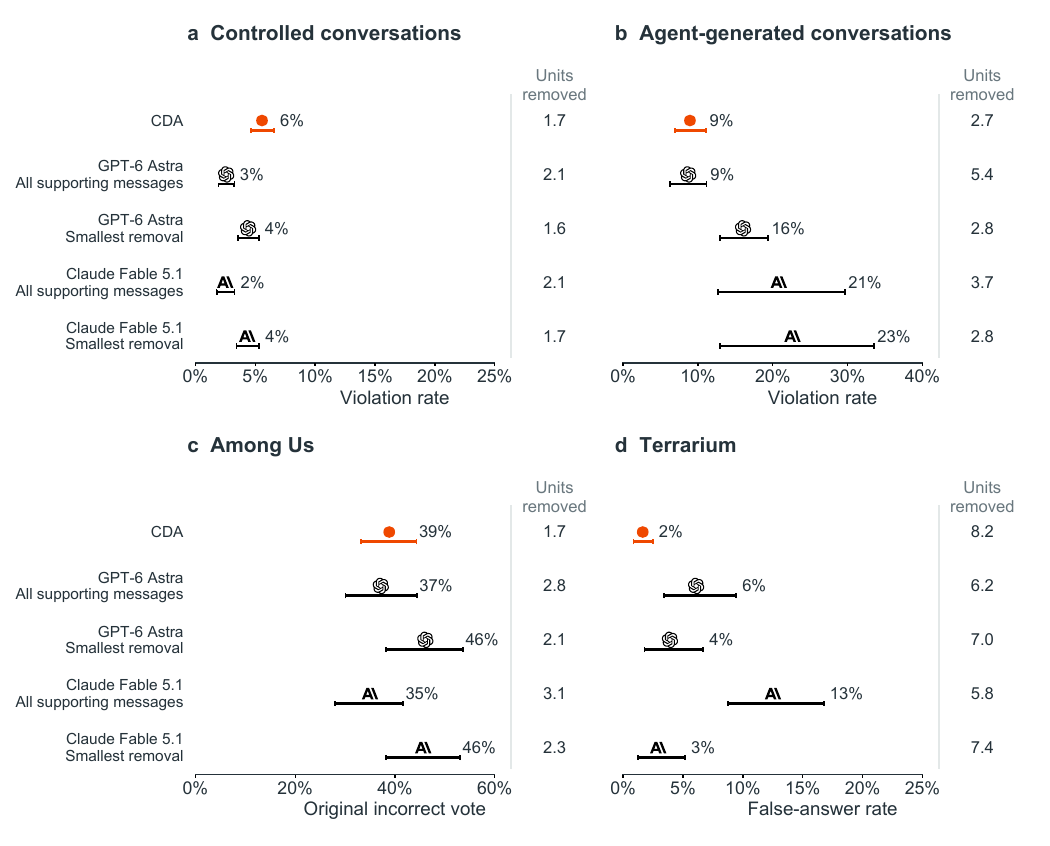}
\caption{\textbf{Frontier monitors differ in intervention size, effectiveness, and coverage.} Points show target-response rates, bars below them show 95\% cluster-bootstrap intervals, and adjacent values give mean units removed. Axis ranges differ. CDA covers 1,670 controlled conversations, 485 agent-generated violations, 130 incorrect votes, and 220 false-belief cases. Claude's controlled and agent-generated rows cover answered subsets of 1,226/1,426 and 75/59 cases, respectively, for attribution/smallest-removal prompts. Among Us smallest-removal rows cover 125 Astra and 129 Claude cases; Terrarium monitor rows cover 215 nonempty selections. Refusals and empty selections receive no assumed intervention outcome. Matched-case comparisons appear in Supplementary Note~\ref{note:comparisons}.}
\label{fig:cross-baselines}
\end{figure}

Comparisons across the four settings assessed both the remaining unwanted behaviour and the number of messages removed (Fig.~\ref{fig:cross-baselines}).

In agent-generated operational conversations, CDA matched Astra's violation reduction with half as many deletions. Astra's attribution selection left 9\% [6, 11] violations with 5.4 messages removed, compared with CDA's 9\% after removing 2.7, although Astra restored task success more often, at 89\% against 79\%. Asked for a smallest removal, Astra selected 2.8 messages and left 16\% [13, 19] violations. Claude refused 85\% of attribution requests and 88\% of smallest-removal requests. On its answered cases, these selections left 21\% and 23\% violations, compared with CDA's 10\% and 13\% on the same cases. Supplementary Note~\ref{note:unplanted} reports intervals and matched-case comparisons.

External settings further link deletion size and effectiveness. In Among Us, Astra's and Claude's attribution selections left 37\% and 35\% harmful votes, removing 2.8 and 3.1 units against CDA's 39\% with 1.7. In Terrarium, their smallest-removal selections left 4\% and 3\% false answers after 7.0 and 7.4 deletions, approaching CDA's 2\% after 8.2 deletions. Prompt-specific outcomes and coverage appear in Supplementary Note~\ref{note:comparisons}.

Policy judgement and prediction of an agent's behaviour required different information. Matched detection prompts assessed harmful conversations against decoys whose mismatched names disconnected the supporting facts, as when logs went to Kestrel but maintenance concerned Harrow. For this distinction, Gemini and Astra reached AUROC 0.98 and 0.93 with no decoy false alarms, compared with 0.73 for CDA's full-context action probability $p_{\mathrm{full}}$. For predicting whether the actor actually acted, $p_{\mathrm{full}}$ reached 0.98 and both monitors 0.68, with monitor scores near chance within harmful conversations or decoys separately. CDA takes the action class as given and assesses its supporting mechanism, while monitors supply the semantic policy judgement (Supplementary Note~\ref{note:comparisons}).

Population-level estimates retain unchanged outcomes wherever CDA supplied no intervention, showing how coverage limited its overall effect (Supplementary Note~\ref{note:comparisons}).

\section*{Discussion}\label{sec:discussion}

Individually aligned agents can nevertheless produce misaligned collective behaviour~\citep{shen2026organizations}. As we enter an agentic era of AI, our focus on AI safety needs to expand from single agent evaluations, which remain equally important, to inter-agent coordination and communication. This becomes all the more pressing when we consider that Cloudflare observed more than fifteenfold growth in user-triggered AI web crawling during 2025~\citep{belson2025radar} while state-linked actors have used AI for covert influence campaigns~\citep{watts2024iran} and agent-orchestrated cyberattacks targeting financial institutions, chemical manufacturers, and government agencies~\citep{anthropic2025espionage,mitre2026campaign}. Indeed, agents performing timed web-research tasks have already used public websites to exchange answers despite restrictions on internet writing, raising further concerns about agents whose operators explicitly encourage harmful coordination~\citep{vonarx2026wiki}. 

This article has introduced CDA which has sought to make messages supporting an action (policy violating or otherwise) amenable to both explanation and intervention. By extending the concepts of necessity and sufficiency and input-removal methods~\citep{watson2021necessity,carter2019subsets,covert2021removing}, CDA is set to identify sufficient combinations of implicated messages and select the smallest support-interrupting removal among tested sets. This complements a burgeoning literature of counterfactual analysis of agent trajectories and semantic support relations~\citep{shah2026replay,bonagiri2026causalflow,jiang2026semantic,goodall2026information}. Across controlled violations, ``natural'' agent-generated interactions, a game, and hosted models, retesting the deciding agents following CDA (and comparator) evaluations has facilitated assessments of behavioural effects. Among qualifying controlled and agent-generated conversations, CDA's removals reduced violations from 93\% to 6\% and from 98\% to 9\%, respectively, and reduced false answers from 58\% to 2\% in the original hosted agents. In short, CDA proffers an explanation that can subsequently propose a concrete intervention whose consequences can be measured.

Nonetheless, deployment must account for the costs of an incorrect intervention. Removals disrupted correct game votes and both beneficial and regretted task choices, and the request task produced no violations. Restricted search, incomplete joint-group recovery, sampled-score instability, and quadratic token-processing costs limit coverage and scale. Explanations remain conditional on the reader and context partition. Three frontier models yielded majority agreement of 0.96 with operational labels ($\kappa=0.91$) and judged 0.94 of explanations plausible or partly plausible, while human validation remains outstanding. These boundaries define the next deployment tests, including sustained interactions and preservation of legitimate outcomes.

Redundant information presents a particularly concerning collective failure mode, particularly as it is resistant to single-message removal. Indeed, redundancy predominated among qualifying agent-generated violations and Terrarium false beliefs, allowing harmful behaviour to survive single-message removal. In a notable example, agents performing web-research tasks also circumvented restrictions on internet writing to share answers through public wikis~\citep{vonarx2026wiki}, where subsequent analysis identified copying between agents as a principal driver of their collective behaviour~\citep{demarzo2026copying}. Though this was the most common collective dependence mechanism in the agent-native interactions condition, joint mechanisms may be at play (or have higher propensity to appear) in different multi-agent settings. Assessing different network and hierarchical organisations of agents is an important next avenue to examine the prevalence of such mechanisms.

In practice, and in order to maximise conversational oversight and minimise chances of emergent misalignment amongst agents, CDA could be combined with semantic and activation-based monitoring and task-preservation checks~\citep{rose2026collusion,tonini2026arbiter,greenblatt2024control}. Indeed, frontier monitors recognised policy-relevant distinctions and selected effective removals, while CDA supplied dependence structure and a prospective warning for an operator-specified action. Its operational interventions matched Astra's all-support selection with half the deletions, although Astra restored legitimate performance more often. Testing that integrated system against adaptive adversaries is an essential next step.

As agents gain operational authority, CDA provides a way to test which shared information sustains a dangerous decision and where intervention can interrupt it.

\section*{Methods}\label{sec:methods}
\subsection*{Decision scores and message interventions}\label{sec:decision-scores}
Each decision has a fixed background containing instructions and task information, and a candidate message set $M$ containing $N$ removable units. A retained subset $S\subseteq M$ determines the reader input. For the planted and unplanted tasks, the reader is Llama-3.3-70B-Instruct in the four-bit bitsandbytes checkpoint \texttt{unsloth/Llama-3.3-70B-Instruct-bnb-4bit} with the output head evaluated in float32, and every other Llama-3.3-70B reader or locally served actor in this article uses the same checkpoint. The second reader in the planted tasks is \texttt{Qwen/Qwen2.5-32B-Instruct} in bfloat16 with the same float32 head. The reader input places the fixed instructions and the rendered context in the model's chat template, followed by a fixed decision prefix that ends immediately before the token committing the response, and the next-token distribution at that position is evaluated over the full vocabulary. Each response class is a short list of surface forms for one response, such as the two task answers of a planted board, the approval and denial variants of the request task, or the on\_trip and no\_trip answers in Terrarium, and its probability mass is the sum of next-token probabilities over the distinct first tokens of those forms, with the two token sets required to be disjoint. If $q_a(S)$ and $q_b(S)$ are these masses for the target and comparison classes, the raw score is $g(S)=\log_2[q_a(S)/q_b(S)]$. The conditional target probability is $q_a(S)/(q_a(S)+q_b(S))$, and excluded mass is $1-q_a(S)-q_b(S)$. Intact contexts with excluded mass above 0.5 receive a separate flag. Conditional class probabilities and sampled action frequencies are distinct measurements.

CDA uses $h(S)=\max\{0,\min\{g(S),c\}\}$, with $c=4$ bits, for its profiles, search, and subset tests so as to bound the quantity between 0 and 4. Clipping discards variation outside the chosen range and can change numerical interactions, so mechanism labels refer to this scale. Define $D=h(M)-h(\varnothing)$, necessity $\delta_i=h(M)-h(M\setminus\{i\})$, and sufficiency $s_i=h(\{i\})-h(\varnothing)$. Analysis requires $D>0$, and planted-reference recovery additionally requires conditional target probability above one half. Non-eligible inputs remain in the reported denominator tables. Physical deletion removes the candidate's text and author tag, with surviving posts in their original order. The fixed background remains present, and the empty board uses the renderer's empty-state text.

\subsection*{Search, subset tests, and classification}\label{sec:search}
The removal search begins with $R=M$ and visits messages in descending $\max\{\delta_i,s_i\}$, breaking ties by input order. Restore message $i$ whenever $h(M\setminus(R\setminus\{i\}))\leq h(\varnothing)+(1-\theta)D$, and repeat once over remaining members. Here $\theta=0.7$ specifies the fraction of added support to remove. The restoration search begins with no retained messages, adds messages in descending profile order until their support above $h(\varnothing)$ reaches $\theta D$, and then removes each added message in reverse order whenever sufficiency persists. Its tie-breakers are larger raw deletion effect and then input order. Denote its final retained set by $S_{\mathrm r}$. Under monotone support, the first removal pass and the pruned restoration set are inclusion-minimal for their respective criteria, without a general minimum-size guarantee. Non-monotone scores can invalidate inclusion-minimality.

Let $r_i=g(M)-g(M\setminus\{i\})$ denote the raw deletion effect, $m=\operatorname{median}_i r_i$, and $a=\operatorname{median}_i|r_i-m|$. Set $a_*=a$ when $a>0$ and $a_*=0.5$ bits otherwise. With individual threshold $\tau=0.3$ and null calibration $k(N)$ defined below, the candidate set is
\begin{equation}
C=R\cup\{i:s_i>\tau D\}\cup\{i:\delta_i>\tau D\ \text{and}\ (N<6\ \text{or}\ r_i>m+k(N)a_*\ \text{or}\ i\in S_{\mathrm r})\}.
\end{equation}
A necessary message qualifies if its raw drop clears the floor or the restoration search includes it in a sufficient set. The raw-drop screen applies only to necessity-based admission, with restoration membership providing corroboration for that route, and the procedure omits the screen below six units. Rank candidates by $\max\{\delta_i,s_i\}$, prioritising $R$ on ties and then input order. The exact core $K$ contains at most the highest eight candidates, and $C\setminus K$ remains explicitly unresolved.

Enumerate every subset of $K$ while keeping the surrounding messages $B=M\setminus K$ visible. Define the core's support $D_K=h(M)-h(B)$ and the qualifying sufficient sets and cuts
\begin{align}
\mathcal S&=\{T\subseteq K:T\ne\varnothing,\ h(B\cup T)-h(B)\geq\theta D_K\},\\
\mathcal C&=\{U\subseteq K:U\ne\varnothing,\ h(M)-h(M\setminus U)\geq\theta D\}.
\end{align}
Their inclusion-minimal members form $\mathcal S_{\min}$ and $\mathcal C_{\min}$. The certified explanatory group is
\begin{equation}
G=\bigcup_{T\in\mathcal S_{\min}}T\ \cup\!\bigcup_{U\in\mathcal C_{\min}}U.
\end{equation}
When this union is empty, the implementation returns $K$ as an uncertified fallback. The intervention $I$ is a smallest member of $\mathcal C_{\min}$, breaking ties by larger score drop and then input order, with $R$ as fallback if no cut exists within $K$. Sufficiency relative to $B$ requires positive $D_K$ for interpretation. Certification applies to the scored core under the fixed background and thresholds, without establishing global completeness or a behavioural guarantee.

Classification checks the following rules in order. A singleton in both minimal collections gives the single label. At least two minimal sufficient sets, all singletons, give redundancy. One multi-message minimal sufficient set with a proper subset that is a cut gives joint dependence. If that sufficient set is also the sole minimal cut, the implementation assigns additive, a graded-support label that does not establish exact additivity. Other patterns are mixed, and absent sufficient sets give none. Supplementary Note~\ref{note:scope} gives the scope of the score and the sensitivity analysis.

\subsection*{Null calibration and threshold transfer}\label{sec:calibration}
The necessity floor calibrates the largest routine deletion effect on harmless conversations of the same size. For each harmless twin board $b$ with $N$ messages, let $J_b$ contain its non-planted posts, let $r_{b,i}$ denote its raw deletion effect for post $i$, and let $m_b$ and $a_{*,b}$ denote the median raw drop and its absolute-deviation scale, calculated over all posts as above. Define
\begin{equation}
z_b=\max_{i\in J_b}\frac{r_{b,i}-m_b}{a_{*,b}},\qquad k(N)=z_{(\lceil0.95n_N\rceil)},
\end{equation}
where $n_N$ is the number of harmless twins at size $N$ and $z_{(j)}$ denotes the $j$th smallest board maximum. With fewer than twenty twins this rule selects the maximum. The raw necessity floor on a new board is $m+k(N)a_*$. This calibration targets approximately one benign board in twenty with any bystander exceeding the floor under a matched null distribution. It does not calibrate the entire candidate set, which also admits removal-search and sufficiency selections and corroborates necessary restoration members, or guarantee an error rate after transfer to different contexts.

The Llama calibration gives $k(N)$ values of 6.04, 9.29, 11.76, 10.52, 9.40, and 5.28 at ten, thirty, fifty, one hundred, two hundred, and five hundred messages, using 840, 840, 840, 140, 70, and 9 harmless twins. Between adjacent calibrated sizes, interpolate linearly in $\log N$, clamping beyond the endpoints. This gives 7.43 at sixteen units in the unplanted tasks and 6.04 at six units in Among Us. The Qwen reader uses its own harmless-twin calibration, giving 5.75, 10.80, and 12.18 at ten, thirty, and fifty messages. The harmless pool includes counterparts from all four task families. Access and payments are held out from support-threshold tuning, while their harmless counterparts contribute to null calibration.

The support settings are $\theta=0.7$, $\tau=0.3$, clipping to $[0,4]$ bits, and an eight-message exact core. A development-only sensitivity analysis evaluates the 36 combinations of $\theta\in\{0.6,0.7,0.8\}$, $\tau\in\{0.2,0.3,0.4,0.5\}$, and $c\in\{2,4,8\}$ bits on 990 eligible harmful audit and backup boards, with null calibration throughout. These settings remain fixed in the held-out and external evaluations. Four bits correspond to conditional target probability $16/17$, linking the clipping choice to near-saturation of action support. For application to another system, estimate the null floor from matched harmless decisions, check the clip against probability saturation, and validate support thresholds through replay. Supplementary Note~\ref{note:scope} reports sensitivity and the limits of this calibration.

\subsection*{Controlled evaluation and behavioural references}\label{sec:controlled-evaluation}
The factorial design contains four task families, seven arrangements, three message counts, three valences, and thirty seeds, giving 7,560 inputs. Audit and backup tasks supply development data, while access and payments supply held-out evaluations at fixed settings. The valences are harmful, permissible counterpart, and mismatched-name decoy. Twenty sampled continuations per replay arm at temperature 0.7, nucleus probability 0.95, and a 120-token generation cap estimate structured-action rates, with 55,069 replay records covering named interventions and member-retention combinations and 3,054 additional comparator records. The eligible comparator population contains 1,670 harmful boards, including 990 development and 680 held-out boards, and uses matching permissible counterparts for task-preservation checks.

The scaling evaluation uses the development audit and backup families, with 140 harmful boards at $N=100$ and 70 at $N=200$, together with harmless twins at each size. The $N=500$ pilot contains nine harmful audit boards spanning single, redundant-triple, and mixed arrangements, with nine harmless twins. Recovery conditions on score eligibility, while scaling replay rates include every harmful board at the corresponding size. These tests measure transfer across context size within development families. Supplementary Note~\ref{note:scaling} and Supplementary Fig.~\ref{fig:si-scaling} report recovery, behavioural outcomes, model-evaluation counts, and timings on one H100.

The behavioural reference enumerates retained subsets of planted slots and marks a subset as acting when at least half its continuations violate the policy. Minimal sufficient sets and cuts follow from these binary outcomes, with cuts requiring the retained complement to fall below one half. Recovery against this reference requires an intact rate of at least one half and remains conditional on the planted-slot partition. Planted-reference recovery instead uses the score-based eligibility rule. Named replay arms retain the full context or remove the selected intervention, explanatory group, strongest individual message, a size-matched random set, or the judge-named set. Legitimate-task success accompanies violation rates. Comparators include rankings of individual removals, sampled Shapley allocations at a matched evaluation budget, and full-transcript judges. The first two receive the planted group size for group recovery, while the main frontier prompt asks judges to identify all relevant posts without supplying that size. The attribution prompt requests all contributing messages and a binary multiple-message judgement. A separate prompt requests a smallest intervention.

Matched detection prompts gave the three frontier monitors the same policy, runbook, and board as the actor, without asserting that a violation occurred, and requested a 0--100 probability that the posts would lead to the forbidden setting. Each model assessed 2,520 harmful boards, 2,520 mismatched-name decoys, and 2,520 permissible counterparts. The normative contrast separated harmful boards from decoys, whose facts did not supply a connected basis for the act. The behavioural contrast classified the 5,040 harmful and decoy boards by whether intact replay selected the action in at least half of samples, yielding 2,299 acting and 2,741 non-acting boards. CDA's comparator was $p_{\mathrm{full}}=q_a(M)/(q_a(M)+q_b(M))$, computed by the actor's own weights, so agreement with its behaviour is partly by construction. AUROC intervals resampled boards, and thresholded detection used probability 0.5. Fable refusals counted as maximal warnings or were excluded, with both outcomes reported in Supplementary Note~\ref{note:comparisons}.

\subsection*{Behavioural replay and causal scope}\label{sec:replay}
Replay resamples the deciding actor under the intact context and specified deletions. A predicted reduction passes when the arm's target-response rate is at most half the intact rate and the two 95\% intervals are disjoint. A predicted limited effect passes when the arm retains at least half the intact rate. Redundancy predicts reductions after removal of $I$ or the explanatory group, and limited effects after the strongest single or random removal. Other mechanism labels predict reductions for the selected, group, and strongest-single arms, with limited effects for random removal. These empirical criteria quantify response patterns and include failed predictions. They do not test exact zero effects. Replay establishes effects of context deletion at the measured decision point, without identifying a unique internal reasoning process or proving every selected message indispensable.

\subsection*{External decision contexts}\label{sec:external-contexts}
The Among Us analysis uses released games in which Llama-3.3-70B plays crewmates against Llama or Phi-4 impostors~\citep{golechha2025amongus}. This analysis extracted 289 harmful votes against crewmates and 719 votes against impostors. Candidate units are up to four observation lines, previous reasoning, and condensed memory. Deletion renumbers observations and resets removed private-state fields to their initial strings, leaving other context fixed. The local actor uses four-bit weights served through vLLM at temperature 0.7, while the released games used an OpenRouter endpoint with unspecified precision. For each retained context, ten reasoning-and-action samples give $x$ votes for the recorded target. Define $\widehat p=(x+0.5)/11$ and $g=\log_2[\widehat p/(1-\widehat p)]$, then apply the same CDA scoring rule. Replay uses a different seed. The analysis covers all 289 harmful cases and all 719 controls, yielding 130 and 266 eligible cases, respectively. Controls comprise 268 cases against Llama impostors and 451 against Phi-4. Thresholds remain fixed from the controlled experiment. GPT-6 Astra receives the exact voter prompt and recorded vote and names the observation lines, previous reasoning, or memory supplying its basis. Fresh replay removes those units. Stored subset samples assess whether a strict subset of an effective judge selection also lowers the vote rate below one half.

Terrarium supplies private meeting-scheduling logs and belief surveys from its authors~\citep{nakamura2025terrarium}. The manuscript describes these records without reproducing their contents. Llama-3.3-70B supplies the external reader, comparing the false travel belief with its alternative. Validity compares 265 labelled believers with 857 exposed non-believers, while CDA dataset contains 265 believers and 975 exposed non-believers. In the persuasion block, the target is the committed task against its best alternative, with positive regret identifying a worse choice. Scores cover 1,615 decisions and CDA records 1,513. Comparison-specific denominators and candidate-set statistics appear in Supplementary Note~\ref{note:terrarium}.

API replay sends the stored system prompt and context to the actor identified in the run configuration, deleting selected messages while retaining the surviving board numbers. Misinformation actors are GPT-4.1 mini, GPT-5.4 mini, and Kimi K2 0905, while persuasion actors are GPT-4.1 mini, GPT-4o, GPT-4o mini, and Kimi K2 0905. Each arm uses ten samples at temperature 0.7, except GPT-5.4 mini, which uses default sampling with low reasoning effort. Its original reasoning setting is unavailable. Kimi replay uses the checkpoint through OpenRouter, so the provider configuration can differ from the original run. Persuasion requests include the original task-selection tool schema, whose returned choices are recorded without execution. Off-format samples count as neither target nor alternative, with their frequency reported. Supplementary Note~\ref{note:terrarium} distinguishes the exported candidate-group replay arm from certified-group removal.

\subsection*{Comparator arms and frontier-judge removals}\label{sec:comparators}
The shared comparison uses intact context and twelve removal arms across all four settings. These remove the selected intervention $I$, the recorded group, the search output $R$, the top $|I|$ units ranked by $\max(\delta_i,s_i)$, the top $|I|$ ranked by $s_i$, the strongest single necessity, a seeded random sample of up to $|I|$ positively scoring units, a uniform random set of $|I|$ visible units, or either of the two selections from each frontier judge. The positive-score pool contains units with $\delta_i>0$ or $s_i>0$ and differs from CDA's thresholded candidate set $C$. Rankings and candidate sampling receive CDA's chosen size and therefore assess selection conditional on that size. Identical removal sets reuse stored replay samples. Planted comparisons additionally include Shapley estimates at CDA's evaluation budget, selecting either the known planted group size or $|I|$, and the second reader's selections on jointly eligible held-out boards. Supplementary Note~\ref{note:comparisons} reports each arm's denominator and selection cost. Reported ranking costs count profile evaluations and exclude the cost of obtaining $|I|$ from CDA.

Both removal-selection prompts use GPT-6 Astra and Claude Fable 5.1 with the actor's decision context and recorded target action. The attribution variant asks for all units supplying information, while the intervention variant asks for a smallest set whose deletion would prevent the action. The Terrarium intervention prompt explicitly mentions removing all copies of repeated information. Its context includes the actor's system prompt and turn up to the survey block, together with the target question and recorded belief. Each of the four Terrarium judge arms has nonempty selections for 215 of 220 eligible believers, with no removal replay for empty selections. Astra unplanted arms cover all 485 eligible violations, while Claude arms cover 75 attribution and 59 smallest-removal responses. Claude planted arms cover 1,226 and 1,426 eligible boards, respectively. The Astra Among Us attribution arm covers all 130 eligible harmful votes and its smallest-removal arm 125, omitting five empty selections. Completed judge selections undergo the same actor replay as other arms in their experiment. Claude answers both Terrarium prompts without refusal. Claude Among Us attribution replay covers all 130 eligible harmful votes and smallest-removal replay covers 129. Both attribution arms cover all 266 eligible controls, while smallest-removal arms cover 234 controls for Astra and 265 for Claude. Claude refuses one harmful and one control smallest-removal request, with no attribution refusals. Other game-control arms cover all 266 eligible cases. Matched-case comparisons restrict each alternative to the cases Claude answered, while population-level policy estimates retain intact outcomes on refusals. Refusals are reported separately from empty selections and behavioural replay outcomes. Containment statistics retain the source distinction between candidates $C$ and certified groups $G$.

\subsection*{Policy estimates, case predictions, and sampling stability}\label{sec:policy}
A retrospective intervention policy applies each removal on eligible cases and retains intact replay on ineligible cases or missing selections. Its rate averages over every case in the specified harmful or control population, reporting coverage separately. These estimates do not simulate downstream changes to a whole episode and do not independently identify harmful cases. Removal magnitude reports units and characters relative to the available message context, with the fixed system prompt excluded where applicable. Sender composition identifies distinct message authors and distinguishes the deciding agent's own state from other agents' contributions.

Per-case predictions use the stored subset tests to predict a reduction after strongest-single removal whenever at least one identified minimal cut is a singleton. The categorical comparator predicts reduction for all labels except redundancy. The observed per-case criterion is a replay rate at most half the intact rate, without the disjoint-interval requirement used for pooled mechanism predictions. Planted evaluation additionally restricts to 1,564 eligible boards with intact replay at least one half, while unplanted and harmful game evaluations use 485 and 130 eligible cases. A singleton cut need not identify the top-ranked single message, so agreement is an empirical predictive test. Intervals resample the experiment's clusters.

Among Us stability reruns forty randomly selected eligible harmful cases with two independent ten-sample seeds and one twenty-sample seed per retained context. Eligibility agreement uses all forty, while set and label agreement condition on eligibility in both compared runs. Exact match and Jaccard overlap assess identities, and stored subset samples assess whether alternative selections still meet a behavioural cut criterion. These stability checks do not add independent fresh replay to every repeated selection. Supplementary Note~\ref{note:amongus} reports all comparisons and denominators.

\subsection*{Unplanted decisions and alarm evaluation}\label{sec:unplanted-methods}
Llama-3.3-70B plays every agent in two simulated tasks, with one hundred episodes under each of neutral, incentive, and pressure conditions per task. Service-setting episodes contain eight decisions, while request episodes contain six approvals or rejections subject to a verification requirement. Structured actions and episode facts determine violation labels, with model annotation assessing agreement on a stratified sample. CDA evaluates every violation and a matched non-violating decision, then separately scores every decision context before the action. Settings remain fixed from the controlled experiments.

Replay covers all 503 violations and 503 matched controls, with twenty continuations per distinct retained context from the original Llama actor at temperature 0.7, nucleus probability 0.95, and a 90-token cap. Seeds differ from those used to generate the episodes. Arms retain the intact context or delete $I$, the exported candidate set $C$, the strongest individual necessity, or a uniform random set of $|I|$ units. Identical deletions share samples. The export calls the candidate set \texttt{G}, whereas \texttt{group} stores the certified explanatory group, so the recorded \texttt{minus\_G} arm tests candidate-set removal. Task success compares the parsed rollout decision with the episode's recorded facts. Ineligible cases return empty selected sets, leaving their set-removal arms identical to intact replay. The mechanism-specific pass rules apply to pooled outcomes within each label, with the candidate-set arm taking the group-removal prediction. Supplementary Note~\ref{note:unplanted} reports all outcomes and distinguishes eligibility groups.

Alarm variants filter the full-context action probability by eligibility, a joint, redundant, or mixed label, an explanatory group of at least two messages, or limited single-message dependence. For the last variant define $\sigma=\max_i\min(\delta_i,s_i)/D$ for $D>0$, where $\delta_i$ and $s_i$ are individual necessity and sufficiency. Eligibility together with $\sigma\leq0.5$ excludes cases with a message exceeding half the added support on both measures. This criterion alone does not certify a collective mechanism. Each evaluation ranks eligible decisions by score and selects at most the observed number of violations, with ties broken by identifier. A filter can leave fewer eligible decisions, so precision and recall can differ. Warning coverage counts episodes with an alarm at or before their first violation. This operating-point analysis uses outcome-informed alarm counts. Transferred-threshold analyses instead calibrate on neutral episodes and test incentive and pressure episodes, or use a seeded two-way split of episodes. Calibration chooses the lowest score reaching precision 0.95, falling back to the calibration set's observed-base-rate point if no score reaches that precision. Bootstraps resample calibration and evaluation episodes and re-estimate the threshold. No test-set violation count determines the transferred threshold. These analyses concern the current decision, with no earlier-turn warning (Supplementary Note~\ref{note:unplanted}).

Policy-aware monitors received the deciding agent's system prompt and complete context, with the exact policy sentence used by the outcome scorer quoted in the question. GPT-6 Astra, Claude Fable 5.1, and Gemini 4 Argon received all 4,200 decisions. Prospective prompts withheld the action and requested the probability that the next decision would violate that policy, while post-hoc prompts included the logged call. Monitor probabilities used the same base-rate and transferred-threshold analyses as the alarm. Refusals counted as maximal warnings in one analysis and were excluded in another, with CDA scores recomputed on the answered subset. These comparisons evaluate prediction of an operator-specified action class alongside policy-based judgement of the next decision. Gemini contributed monitor, detection, and annotation assessments, while removal-selection arms used Astra and Fable.

Model annotation covered 200 decisions comprising 100 violations, 50 benign decisions, and 50 uncertain decisions near even odds of the violating action. Each model saw the context, messages, question, action, and CDA explanation, without the reference labels, and judged policy violation and explanation plausibility. Majority violation labels required agreement among answered models, leaving four ties unresolved, while plausibility ties received ``partly''. Agreement with operational labels used 196 majority-labelled decisions, and plausibility used all 200. Fable refused 62 items, including 43 violations. Annotation supplements the structured labels and behavioural tests, with no human judgements in these results (Supplementary Note~\ref{note:unplanted}).

\subsection*{Tracing, computation, and statistics}\label{sec:computation}
One-level tracing scores the observed relay text by summing its token log-probabilities under retained upstream sources, conditioning each token on the preceding observed tokens. Total bits and bits per token describe support for that fixed text, without the action-score clipping. Planted upstream replay deletes the traced sources and regenerates the relay before resampling the downstream actor. In the unplanted worlds, tracing covers 402 own replies with eight-message upstream contexts, while 347 earlier own posts have no upstream messages by construction. Source recovery can include information unrelated to the downstream violation, so unplanted tracing establishes provenance without validating every upstream behavioural link.

Individual profiles and the first removal pass require at most $3N+2$ distinct reader evaluations, with at most $N$ additional evaluations in the second removal pass. Restoration adds at most $N$ evaluations during accumulation and $N$ during pruning. Exact analysis adds at most $2^{|K|}\leq256$, with cached duplicates reused, giving the conservative bound $6N+258$ and linear evaluation count for fixed core size. Each evaluation processes a context whose length grows with the messages in view, so total processed tokens can grow quadratically with $N$ at fixed mean message length. Model attention, prefix caching, and sampled generation further determine runtime. Analyses apply per decision and can run independently across agents. Reported planted timings use one H100.

Confidence intervals use percentile cluster bootstraps, with 2,000 resamples of boards for planted results, games for Among Us, and runs for Terrarium, 2,000 episode resamples for unplanted replay, and 1,000 for unplanted alarms and for the planted detection tables, which resample harmful and decoy pairs. Intervals are 95\%, and sampled continuations within a decision are not independent experimental replicates. Tables report the case denominators for each eligible subset. Recovery, response frequencies, and reader discrimination have distinct denominators and interpretations. Model-annotation intervals use 2,000 decision-level bootstrap resamples, with agreement and Cohen's $\kappa$ calculated against operational labels on answered items.

\section*{Data availability}\label{sec:data}
Data are available upon reasonable request. Among Us analyses use the CC0 game records released by Golechha and Garriga-Alonso~\citep{golechha2025amongus}. Terrarium interaction logs were supplied by the authors of the source study~\citep{nakamura2025terrarium} and are not redistributed.

\section*{Code availability}\label{sec:code}
Code is available upon reasonable request.

\section*{Acknowledgements}\label{sec:acknowledgements}
The author thanks Felix P. S. Wallis for support and discussion during this work and the authors of the Terrarium study for supplying their interaction logs. Artificial Societies funded this work.

\section*{Author contributions}\label{sec:contributions}
E.C. conceived the study, developed the methodology and software, conducted the experiments, analysed the results, prepared the figures, and wrote and revised the manuscript.

\section*{Competing interests}\label{sec:interests}
E.C. is an employee of Artificial Societies and holds equity in the company.

\clearpage
\section*{Supplementary information}\label{sec:supplementary}
\newcounter{suppnote}
\addcontentsline{toc}{section}{Supplementary information}
\renewcommand{\thetable}{S\arabic{table}}
\renewcommand{\thefigure}{S\arabic{figure}}
\renewcommand{\theequation}{S\arabic{equation}}
\setcounter{table}{0}
\setcounter{figure}{0}
\setcounter{equation}{0}
\renewcommand{\theHequation}{S\arabic{equation}}
\renewcommand{\theHfigure}{S\arabic{figure}}
\renewcommand{\theHtable}{S\arabic{table}}

\refstepcounter{suppnote}
\edef\notesection{\noexpand\section*{Supplementary Note \thesuppnote. Score interpretation, calibration, and scope}}
\notesection\label{note:scope}
CDA operates on a specified decision, response contrast, and partition of removable context. For a retained message set $S$, the reader assigns target and comparison masses $q_a(S)$ and $q_b(S)$, giving $g(S)=\log_2[q_a(S)/q_b(S)]$ and bounded support $h(S)=\max\{0,\min\{g(S),c\}\}$. The probability conditional on the two classes is $q_a/(q_a+q_b)=1/(1+2^{-g})$. This quantity equals the unconditional target mass only when the two classes exhaust the output distribution. Excluded mass $1-q_a-q_b$ therefore accompanies the scores, with intact contexts above 0.5 flagged separately. Sampled action frequencies provide a distinct behavioural measurement.

The fixed background remains visible in every evaluation, while $M$ denotes the $N$ removable messages. Added support is $D=h(M)-h(\varnothing)$, individual necessity is $\delta_i=h(M)-h(M\setminus\{i\})$, and individual sufficiency is $s_i=h(\{i\})-h(\varnothing)$. Positive necessity indicates that deletion reduces support, and negative necessity indicates that deletion increases it. Sufficiency measures support above the fixed background when retaining one message. Clipping defines the scale of these quantities and the resulting labels, so neither profile is a logical assertion about all possible contexts.

Null calibration screens admission through individual necessity. Let $r_i=g(M)-g(M\setminus\{i\})$, $m=\operatorname{median}_i r_i$, and $a=\operatorname{median}_i|r_i-m|$. The implementation uses $a_*=a$ for positive $a$ and $a_*=0.5$ bits when $a=0$. Each harmless twin contributes the largest $(r_i-m)/a_*$ over its non-planted posts, with the median and deviation calculated over every post. The empirical 0.95 quantile of these board maxima defines $k(N)$, using order statistic $\lceil0.95n_N\rceil$ for $n_N$ twins. A necessity candidate must satisfy $\delta_i>\tau D$ and either exceed $m+k(N)a_*$ in raw drop or belong to the restoration set $S_{\mathrm r}$ defined below. Removal-search members and sufficiently strong individual-retention scores enter independently, and the floor does not apply below six units. The null criterion concerns the raw-drop route under matched benign conditions, without establishing a false-positive rate for the whole candidate set or transferred contexts.

\begingroup\small
\setlength{\tabcolsep}{3pt}
\begin{longtable}{>{\raggedright\arraybackslash}p{3.66cm}>{\raggedright\arraybackslash}p{3.66cm}>{\raggedright\arraybackslash}p{3.66cm}>{\raggedright\arraybackslash}p{3.66cm}}
\caption{Reader-specific null calibration. Harmless counterparts from all four main task families contribute at ten, thirty, and fifty messages. Larger-size calibration uses the development-family scaling design.}\\
\toprule
Messages & Llama twins & Llama $k(N)$ & Qwen $k(N)$\\
\midrule\endfirsthead
\toprule
Messages & Llama twins & Llama $k(N)$ & Qwen $k(N)$\\
\midrule\endhead
10 & 840 & 6.04 & 5.75\\
30 & 840 & 9.29 & 10.8\\
50 & 840 & 11.76 & 12.18\\
100 & 140 & 10.52 & --\\
200 & 70 & 9.4 & --\\
500 & 9 & 5.28 & --\\
\bottomrule\end{longtable}
\endgroup

For an intermediate size $N$ between calibrated sizes $N_j$ and $N_{j+1}$, the transfer rule is
\begin{equation}
k(N)=k(N_j)+\frac{\log N-\log N_j}{\log N_{j+1}-\log N_j}\bigl[k(N_{j+1})-k(N_j)\bigr].
\end{equation}
Outside the calibrated range, the nearest endpoint applies. The Llama floor therefore uses $k(16)=7.43$ in operational worlds and $k(6)=6.04$ in the game. External transfer uses each decision's number of removable units, with no external outcome labels used to choose this function. Distribution shift can change the null behaviour, and the nine-twin calibration at 500 messages provides especially limited precision.

The removal search maintains $h(M\setminus R)\leq h(\varnothing)+(1-\theta)D$. It starts at $R=M$ and tries restoring each message in descending $\max(\delta_i,s_i)$, with input order breaking ties. A successful restoration preserves the inequality and removes that message from $R$. Under monotone support, a restoration that fails cannot become acceptable after further restorations, so the first pass leaves an inclusion-minimal removal set. This proof does not establish smallest cardinality. Non-monotone scores can invalidate that implication, so the implementation performs a second pass and records changes.

The restoration search finds a sufficient retained set under the complementary score criterion. Starting with $A=\varnothing$, add units in descending $\max(\delta_i,s_i)$, breaking ties by larger raw deletion effect and then input order, until $h(A)-h(\varnothing)\geq\theta D$. Next, visit the added units in reverse order and remove any unit whose absence preserves this inequality. The resulting set is $S_{\mathrm r}$, with
\begin{equation}
h(S_{\mathrm r})-h(\varnothing)\geq\theta D.
\end{equation}
Under monotone support, a failed pruning deletion cannot become acceptable after further deletions, proving inclusion-minimality. Neither search guarantees smallest cardinality, and non-monotonicity limits both guarantees. In a pure conjunction, each member can form a cut while all members are required for sufficiency. In pure redundancy, every copy suffices alone while interruption requires removing all copies. The two searches therefore supply complementary evidence for the combination tests.

The candidate set combines the removal-search output, sufficient single messages, and necessary messages corroborated by the floor or restoration membership. Ranking by $\max(\delta_i,s_i)$, prioritising removal-search membership on ties and then input order, yields a core $K$ of at most eight messages. Candidates outside $K$ remain unresolved and visible in the surrounding set $B=M\setminus K$. Every subset of $K$ receives a score under this fixed surrounding context. With $D_K=h(M)-h(B)$, define
\begin{align}
\mathcal S&=\{T\in\mathcal P(K)\setminus\{\varnothing\}:h(B\cup T)-h(B)\geq\theta D_K\},\\
\mathcal C&=\{U\in\mathcal P(K)\setminus\{\varnothing\}:h(M)-h(M\setminus U)\geq\theta D\}.
\end{align}
Here $\mathcal P(K)$ is the power set, and $\mathcal S_{\min}$ and $\mathcal C_{\min}$ retain only sets with no qualifying proper subset. The certified group and selected intervention are
\begin{equation}
G=\bigcup_{T\in\mathcal S_{\min}}T\ \cup\!\bigcup_{U\in\mathcal C_{\min}}U,\qquad I\in\underset{U\in\mathcal C_{\min}}{\arg\min}|U|.
\end{equation}
Ties for $I$ favour larger support loss and then input order. If no cut exists within the core, $I=R$ is the fallback. An empty union returns $K$ as an uncertified explanatory fallback. Positive $D_K$ is necessary for interpreting sufficiency above the surrounding context, and certification remains conditional on the scored core and thresholds.

The ordered classification rules distinguish a common singleton in both minimal collections, multiple sufficient singleton routes, one multi-message sufficient set with a proper-subset cut, and a sole multi-message sufficient set that also forms the sole cut. These receive single, redundant, joint, and graded-support labels, respectively. The implementation calls the last label additive, although it establishes no additive score identity. Remaining patterns are mixed, and absent sufficient sets give none. These labels summarise thresholded dependence, with behavioural replay providing the independent test of intervention effects.

Sensitivity evaluates 36 combinations on 990 eligible harmful development boards, with null calibration throughout. The support thresholds remain fixed for the harmful held-out families and external evaluations. The grid changes exact group recovery by at most 1.7 percentage points below the mean-score optimum, label recovery by 2.8, and intervention recovery by 0.9. The grid also contains settings with higher group or intervention recovery individually, at the cost of lower label recovery. This analysis supports robustness within the tested range.

\begingroup\small
\setlength{\tabcolsep}{3pt}
\begin{longtable}{>{\raggedright\arraybackslash}p{4.30cm}>{\raggedright\arraybackslash}p{2.50cm}>{\raggedright\arraybackslash}p{2.50cm}>{\raggedright\arraybackslash}p{2.50cm}>{\raggedright\arraybackslash}p{2.00cm}}
\caption{Development sensitivity. Rows varying one parameter hold the others at the grid optimum. Recovery values are fractions.}\\
\toprule
Settings & Group & Label & Intervention & Unresolved\\
\midrule\endfirsthead
\toprule
Settings & Group & Label & Intervention & Unresolved\\
\midrule\endhead
frozen point ($\theta$ 0.7, $\tau$ 0.3, c 4) & 0.797 & 0.840 & 0.878 & 0.037\\
grid optimum ($\theta$ 0.6, $\tau$ 0.3, c 4) & 0.801 & 0.846 & 0.878 & 0.030\\
$\theta$ 0.6 / 0.7 (frozen) / 0.8 & 0.801 / 0.797 / 0.794 & 0.846 / 0.840 / 0.837 & 0.878 / 0.878 / 0.875 & 0.030 / 0.037 / 0.036\\
$\tau$ 0.2 / 0.3 (frozen) / 0.4 / 0.5 & 0.801 / 0.801 / 0.797 / 0.797 & 0.846 / 0.846 / 0.843 / 0.841 & 0.877 / 0.878 / 0.878 / 0.878 & 0.031 / 0.030 / 0.030 / 0.030\\
c 2 / 4 (frozen) / 8 bits & 0.792 / 0.801 / 0.804 & 0.837 / 0.846 / 0.829 & 0.874 / 0.878 / 0.885 & 0.035 / 0.030 / 0.033\\
\bottomrule\end{longtable}
\endgroup

\begin{figure}[H]
\centering
\includegraphics[width=\linewidth,height=0.73\textheight,keepaspectratio]{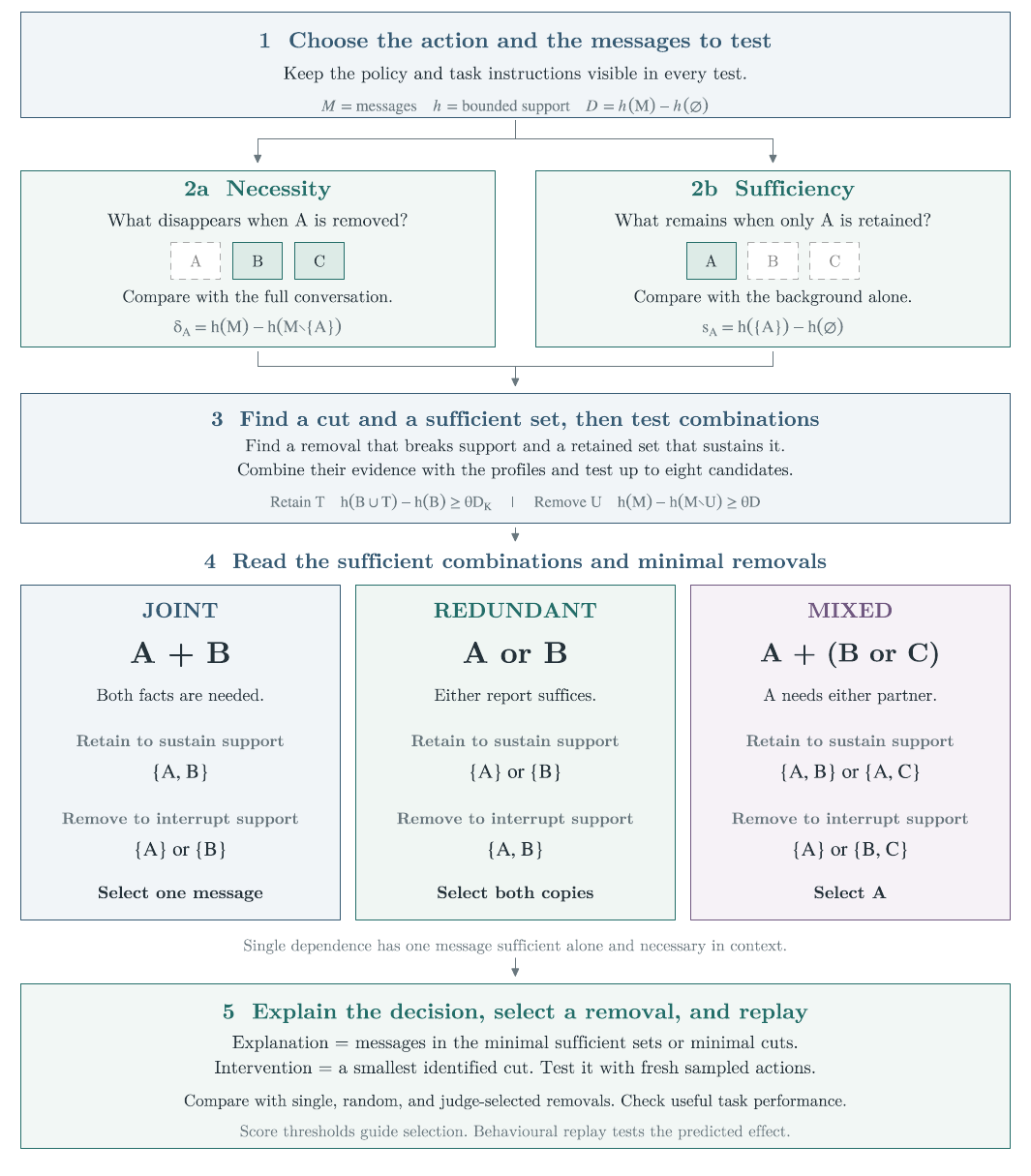}
\caption{\textbf{Detailed workflow.} CDA scores retained message subsets, searches for a removal and a sufficient retained set, and tests all combinations within a bounded core. The notation follows \methodsref{}. Shaded tiles remain visible, and dashed tiles denote deletion. Necessity-based shortlist admission uses the null-calibrated floor or restoration membership, while removal-search and sufficiency admissions remain separate. Joint, redundant, and mixed examples illustrate possible support relationships. Fresh actor replay evaluates the selected intervention.}
\label{fig:si-procedure-detail}
\end{figure}

\refstepcounter{suppnote}
\edef\notesection{\noexpand\section*{Supplementary Note \thesuppnote. Planted mechanisms and behavioural references}}
\notesection\label{note:planted}
The main controlled experiment contains 7,560 inputs from four operational task families, seven supporting arrangements, three conversation sizes, three valences, and thirty seeds. Audit and backup supply development cases, while harmful access and payment cases remain held out from support-threshold tuning. Harmless counterparts from all families supply null calibration. Arrangements comprise one sufficient message, two or three jointly required facts, two or three interchangeable sufficient messages, mixed support, and relayed support. Mixed arrangements require one message together with either of two alternatives. Relayed arrangements place an upstream composition within one downstream post.

Recovery compares the certified group, its mechanism label, and the intervention with the planted reference on 1,670 eligible harmful boards, comprising 990 development and 680 held-out cases. Eligibility requires conditional target probability above one half and positive added support. Top-ranked individual removals and Shapley estimates receive the true planted group size, while the main transcript prompt receives no such size. Score eligibility excludes 34\% of harmful inputs and 83\% of joint triples, including 98\% of held-out joint triples. The four eligible held-out triples give a particularly uncertain recovery estimate.

\begingroup\small
\setlength{\tabcolsep}{3pt}
\begin{longtable}{>{\raggedright\arraybackslash}p{3.20cm}>{\raggedright\arraybackslash}p{5.00cm}>{\raggedright\arraybackslash}p{2.80cm}>{\raggedright\arraybackslash}p{3.00cm}}
\caption{Planted-reference recovery in development families. Group intervals are 95\% board-clustered bootstrap intervals. Label and intervention columns give exact recovery fractions.}\\
\toprule
Mechanism & Group & Label & Intervention\\
\midrule\endfirsthead
\toprule
Mechanism & Group & Label & Intervention\\
\midrule\endhead
single & 0.92 [0.88, 0.96] & 0.94 & 0.94\\
joint2 & 0.77 [0.69, 0.84] & 0.83 & 0.93\\
joint3 & 0.68 [0.56, 0.81] & 0.84 & 0.96\\
redundant2 & 0.87 [0.82, 0.92] & 0.92 & 0.89\\
redundant3 & 0.92 [0.88, 0.96] & 0.99 & 0.92\\
mixed & 0.62 [0.55, 0.69] & 0.61 & 0.78\\
relayed & 0.64 [0.54, 0.72] & 0.69 & 0.75\\
\bottomrule\end{longtable}
\endgroup

\begingroup\small
\setlength{\tabcolsep}{3pt}
\begin{longtable}{>{\raggedright\arraybackslash}p{3.66cm}>{\raggedright\arraybackslash}p{3.66cm}>{\raggedright\arraybackslash}p{3.66cm}>{\raggedright\arraybackslash}p{3.66cm}}
\caption{Group-recovery comparators in development families. Rankings and Shapley receive the true group size. The judge column reports GPT-6 Astra.}\\
\toprule
Mechanism & Single ranking & Shapley & Judge\\
\midrule\endfirsthead
\toprule
Mechanism & Single ranking & Shapley & Judge\\
\midrule\endhead
single & 0.95 & 1.00 & 1.00\\
joint2 & 0.71 & 0.61 & 1.00\\
joint3 & 0.32 & 0.28 & 1.00\\
redundant2 & 0.00 & 0.79 & 1.00\\
redundant3 & 0.01 & 0.37 & 1.00\\
mixed & 0.05 & 0.21 & 1.00\\
relayed & 0.68 & 0.90 & 1.00\\
\bottomrule\end{longtable}
\endgroup

\begingroup\small
\setlength{\tabcolsep}{3pt}
\begin{longtable}{>{\raggedright\arraybackslash}p{3.20cm}>{\raggedright\arraybackslash}p{5.00cm}>{\raggedright\arraybackslash}p{2.80cm}>{\raggedright\arraybackslash}p{3.00cm}}
\caption{Planted-reference recovery in held out families. Group intervals are 95\% board-clustered bootstrap intervals. Label and intervention columns give exact recovery fractions.}\\
\toprule
Mechanism & Group & Label & Intervention\\
\midrule\endfirsthead
\toprule
Mechanism & Group & Label & Intervention\\
\midrule\endhead
single & 0.90 [0.85, 0.96] & 0.93 & 0.94\\
joint2 & 0.62 [0.47, 0.76] & 0.78 & 0.84\\
joint3 & 0.00 (4 boards) & 0.75 & 0.25\\
redundant2 & 0.65 [0.57, 0.72] & 0.49 & 0.54\\
redundant3 & 0.73 [0.66, 0.79] & 0.66 & 0.45\\
mixed & 0.35 [0.25, 0.48] & 0.46 & 0.91\\
relayed & 0.70 [0.61, 0.79] & 0.76 & 0.87\\
\bottomrule\end{longtable}
\endgroup

\begingroup\small
\setlength{\tabcolsep}{3pt}
\begin{longtable}{>{\raggedright\arraybackslash}p{3.66cm}>{\raggedright\arraybackslash}p{3.66cm}>{\raggedright\arraybackslash}p{3.66cm}>{\raggedright\arraybackslash}p{3.66cm}}
\caption{Group-recovery comparators in held out families. Rankings and Shapley receive the true group size. The judge column reports GPT-6 Astra.}\\
\toprule
Mechanism & Single ranking & Shapley & Judge\\
\midrule\endfirsthead
\toprule
Mechanism & Single ranking & Shapley & Judge\\
\midrule\endhead
single & 0.73 & 0.99 & 1.00\\
joint2 & 0.44 & 0.62 & 1.00\\
joint3 & 0.00 & 0.25 & 1.00\\
redundant2 & 0.12 & 0.89 & 1.00\\
redundant3 & 0.01 & 0.52 & 1.00\\
mixed & 0.03 & 0.28 & 1.00\\
relayed & 0.81 & 0.98 & 1.00\\
\bottomrule\end{longtable}
\endgroup

The two searches address complementary support patterns without ensuring complete recovery. A jointly required group can have an effective one-post cut, while its sufficient retained set requires every member. Restoration membership can therefore admit necessary messages whose raw effects fail the null floor. Development mechanism recall is 0.83 for joint pairs and 0.84 for triples, with group recovery of 0.77 and 0.68. Held-out triples receive the correct label in three of four eligible cases and exact group recovery in none. Under deletion, a planted message may also need a bystander to preserve support. In the exhaustive held-out ten-message reference, the planted post alone suffices on sixteen of nineteen single and thirteen of sixteen relayed boards, so restoration can include an additional necessary bystander. These limits distinguish the planted reference from the reader's dependence under deletion.

The behavioural reference enumerates retained subsets of the planted slots using twenty continuations per context. A subset acts when at least half its samples violate the policy, and a cut lowers its retained complement below one half. Planted and behavioural labels agree on 0.85 [0.83, 0.87] of eligible acting boards. Exact group recovery against this behavioural reference reaches 0.89 for single, 0.71 for joint pairs, 0.59 for joint triples, 0.82 and 0.81 for redundant pairs and triples, 0.60 for mixed, and 0.62 for relayed arrangements. This reference remains conditional on the planted-slot partition.

The replay record contains 55,069 named and retained-subset records, with 3,054 additional comparator records. Main mechanism-level replay pools all 360 harmful boards per arrangement, while the cross-comparator analysis conditions on the 1,670 score-eligible cases. Consequently, low intact rates for joint triples and high eligible-population rates describe different denominators. The shared comparator figure and complete controlled figure show these complementary evaluations.

\begin{figure}[H]
\centering
\includegraphics[width=\linewidth,height=0.73\textheight,keepaspectratio]{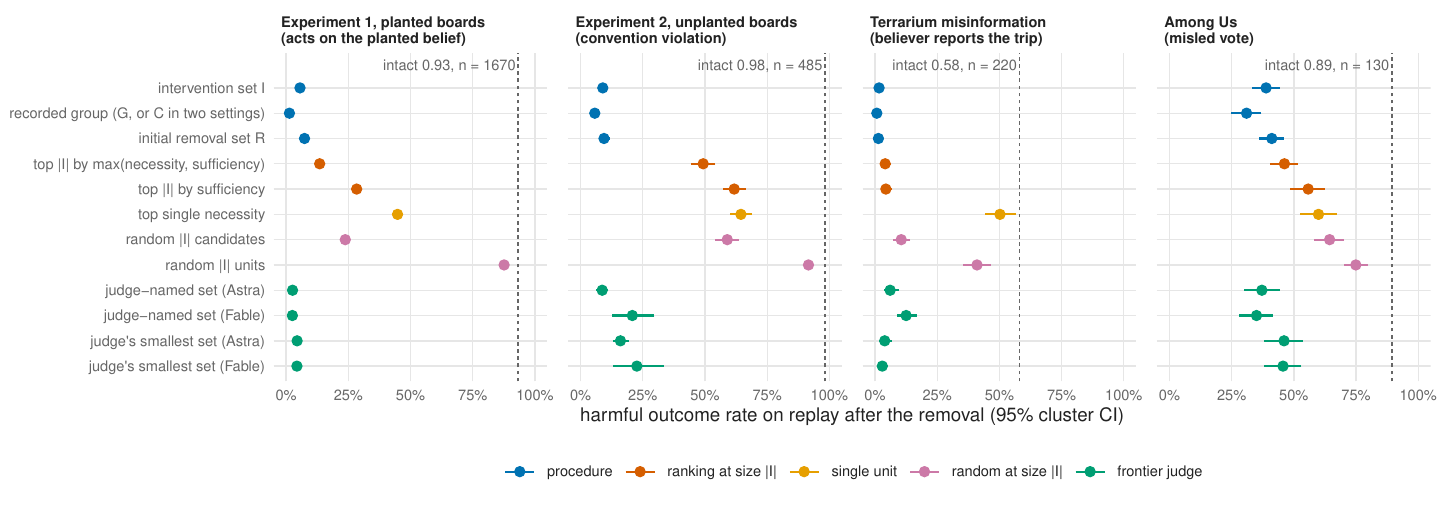}
\caption{\textbf{Shared replay comparisons.} Panels report target responses after each recorded removal in the four settings. Paired changes appear in Note~\ref{note:comparisons}, with selection costs reported in \methodsref{} and Notes~\ref{note:scaling}--\ref{note:unplanted}. Intervals cluster by board, episode, run, or game. Judge arms condition on answered, nonempty selections, as specified in Notes~\ref{note:amongus}--\ref{note:comparisons}. The arm labelled group in the source export denotes candidate-set removal in the unplanted and Terrarium experiments, and certified-group removal in planted tasks and Among Us. Profile rankings receive the selected CDA size. Costs for those rankings exclude obtaining that size, and a judge call differs computationally from an open-weight reader evaluation.}
\label{fig:si-baselines}
\end{figure}

\begin{figure}[H]
\centering
\includegraphics[width=\linewidth,height=0.73\textheight,keepaspectratio]{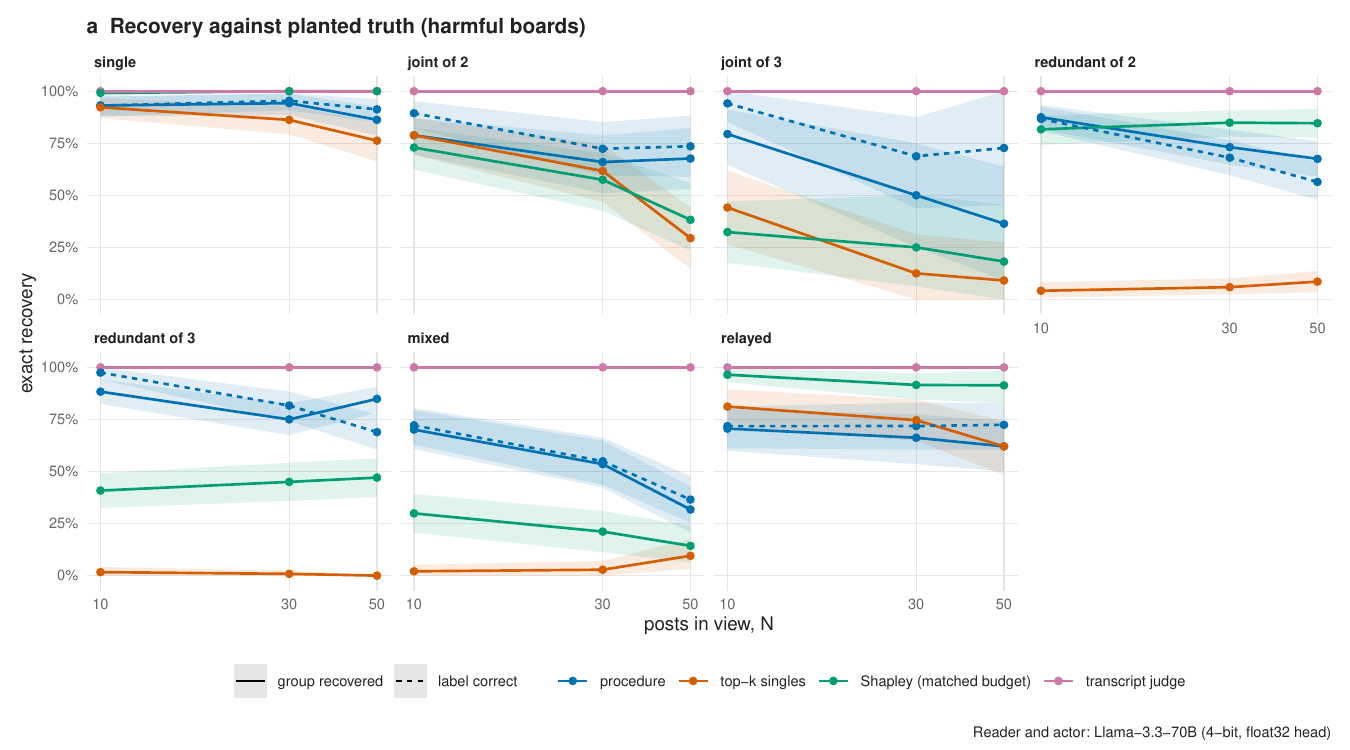}
\caption{\textbf{Controlled recovery against planted truth.} The controlled design tests planted mechanisms at ten, thirty, and fifty messages. Development and held-out family labels refer to support-threshold tuning, with harmless twins from all families contributing to null calibration. Recovery conditions on score eligibility, and reference definitions and denominators appear in this note. Ribbons give board-clustered 95\% intervals.}
\label{fig:si-planted}
\end{figure}

\begin{figure}[H]
\centering
\includegraphics[width=\linewidth,height=0.73\textheight,keepaspectratio]{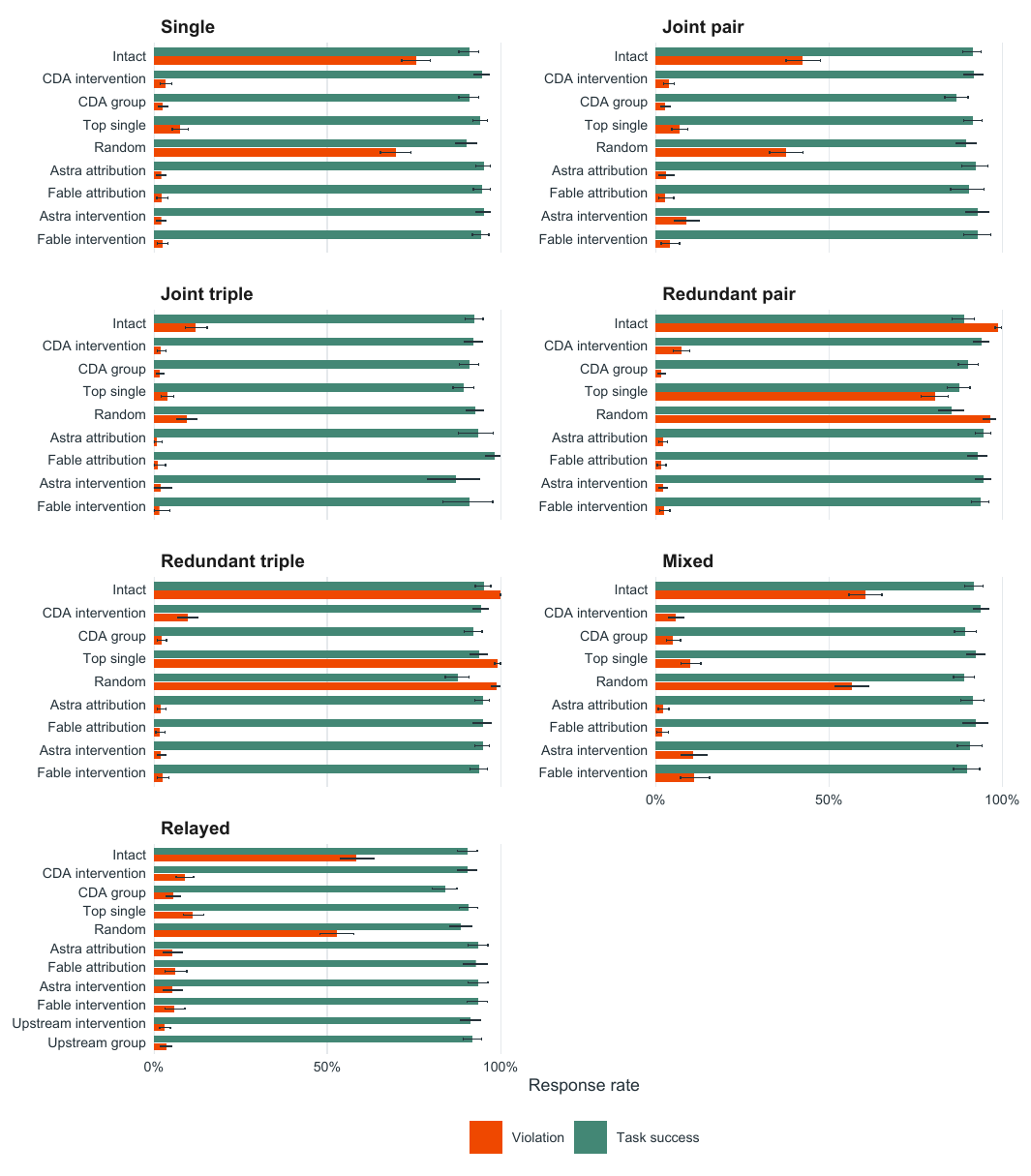}
\caption{\textbf{Behavioural validation and task preservation in controlled conversations.} Bars show violation and task-success rates after each removal. Non-monitor arms include all harmful boards, while monitor arms use eligible boards with returned selections. CDA intervention and group denote removal of $I$ and $G$, and upstream arms remove sources before regenerating a relay. Error bars show 95\% board-clustered bootstrap intervals. Methods specifies the mechanism-level prediction criteria.}
\label{fig:si-planted-replay}
\end{figure}

\begingroup\small
\setlength{\tabcolsep}{3pt}
\begin{longtable}{>{\raggedright\arraybackslash}p{2.40cm}>{\raggedright\arraybackslash}p{1.70cm}>{\raggedright\arraybackslash}p{1.70cm}>{\raggedright\arraybackslash}p{1.70cm}>{\raggedright\arraybackslash}p{1.70cm}>{\raggedright\arraybackslash}p{1.70cm}>{\raggedright\arraybackslash}p{1.70cm}}
\caption{Replay across all harmful planted boards. Each arrangement contains 360 boards, with twenty samples per arm. Entries give violation rates.}\\
\toprule
Mechanism & Intact & Remove $I$ & Remove $G$ & Top single & Random & Astra\\
\midrule\endfirsthead
\toprule
Mechanism & Intact & Remove $I$ & Remove $G$ & Top single & Random & Astra\\
\midrule\endhead
single & 0.76 & 0.03 & 0.03 & 0.07 & 0.70 & 0.02\\
joint2 & 0.43 & 0.04 & 0.03 & 0.07 & 0.38 & 0.02\\
joint3 & 0.12 & 0.02 & 0.02 & 0.04 & 0.09 & 0.00\\
redundant2 & 0.99 & 0.07 & 0.02 & 0.81 & 0.96 & 0.02\\
redundant3 & 1.00 & 0.10 & 0.02 & 0.99 & 0.99 & 0.02\\
mixed & 0.61 & 0.06 & 0.05 & 0.10 & 0.57 & 0.01\\
relayed & 0.58 & 0.09 & 0.06 & 0.11 & 0.53 & 0.03\\
\bottomrule\end{longtable}
\endgroup

Exhaustive scoring of all subsets on 225 eligible ten-message boards tests the restricted search against a complete score reference. Every reported minimal cut satisfies the exhaustive criterion, and the intervention has globally minimum size on all but one case, a fraction rounding to 1.00 [0.99, 1.00]. The restoration set is an exhaustive minimal sufficient set on 224 of 225 cases. The certified group contains a smallest sufficient set on 0.98 [0.96, 1.00], including every eligible joint triple. Additional cuts outside the group occur on 0.52 [0.46, 0.59] of boards. Local certification therefore leaves alternative explanations unresolved.

\begingroup\small
\setlength{\tabcolsep}{3pt}
\begin{longtable}{>{\raggedright\arraybackslash}p{2.40cm}>{\raggedright\arraybackslash}p{1.20cm}>{\raggedright\arraybackslash}p{2.10cm}>{\raggedright\arraybackslash}p{2.10cm}>{\raggedright\arraybackslash}p{3.50cm}>{\raggedright\arraybackslash}p{2.40cm}}
\caption{Exhaustive ten-message reference. Counts condition on score eligibility, and rates concern the full subset-score reference.}\\
\toprule
Mechanism & Cases & Cuts exact & Smallest $I$ & Sufficient set in $G$ & Extra cuts\\
\midrule\endfirsthead
\toprule
Mechanism & Cases & Cuts exact & Smallest $I$ & Sufficient set in $G$ & Extra cuts\\
\midrule\endhead
single & 38 & 1.00 & 0.97 & 1.00 & 0.34\\
joint2 & 28 & 1.00 & 1.00 & 0.96 & 0.54\\
joint3 & 13 & 1.00 & 1.00 & 1.00 & 0.85\\
redundant2 & 40 & 1.00 & 1.00 & 1.00 & 0.35\\
redundant3 & 40 & 1.00 & 1.00 & 1.00 & 0.42\\
mixed & 34 & 1.00 & 1.00 & 0.97 & 0.65\\
relayed & 32 & 1.00 & 1.00 & 0.94 & 0.81\\
all & 225 & 1.00 & 1.00 & 0.98 & 0.52\\
\bottomrule\end{longtable}
\endgroup

Permissible twins quantify a specificity cost. Applying removals selected from their harmful counterparts lowers task success from 0.87 [0.85, 0.88] to 0.83 after $I$ and 0.78 after $G$. Shapley selections supplied with the planted size leave 0.77, while search, rankings, and judge selections leave 0.82--0.84. Effective removal of harmful support can also remove information needed for a legitimate counterpart.

\refstepcounter{suppnote}
\edef\notesection{\noexpand\section*{Supplementary Note \thesuppnote. Computational cost and longer conversations}}
\notesection\label{note:scaling}
The individual profiles require at most $2N+2$ scores, and the first removal pass adds at most $N$. The second removal pass adds at most $N$. Restoration requires at most $N$ additions and at most $N$ pruning tests. The exact core adds at most $2^{|K|}\leq256$, with cached duplicates reused, giving the conservative bound $6N+258$. This count is linear in message count for a fixed core cap. Total processed tokens can grow quadratically because each evaluation reads a context whose length grows with $N$, and model attention, caching, and sampled generation further determine runtime. In the main planted run, restoration adds means of only 0.3, 0.5, and 0.6 uncached evaluations at ten, thirty, and fifty messages, because profiles often already score its retained sets.

\begingroup\small
\setlength{\tabcolsep}{3pt}
\begin{longtable}{>{\raggedright\arraybackslash}p{3.66cm}>{\raggedright\arraybackslash}p{3.66cm}>{\raggedright\arraybackslash}p{3.66cm}>{\raggedright\arraybackslash}p{3.66cm}}
\caption{Measured main-run cost per board on one H100, pooled over families and valences.}\\
\toprule
Messages & Evaluations & Seconds & Reader tokens\\
\midrule\endfirsthead
\toprule
Messages & Evaluations & Seconds & Reader tokens\\
\midrule\endhead
10 & 39.0 [37.4, 40.8] & 7 & 12,000\\
30 & 125.3 [120.6, 129.9] & 31 & 62,000\\
50 & 206.3 [198.3, 214.5] & 63 & 143,000\\
\bottomrule\end{longtable}
\endgroup

\begin{figure}[H]
\centering
\includegraphics[width=\linewidth]{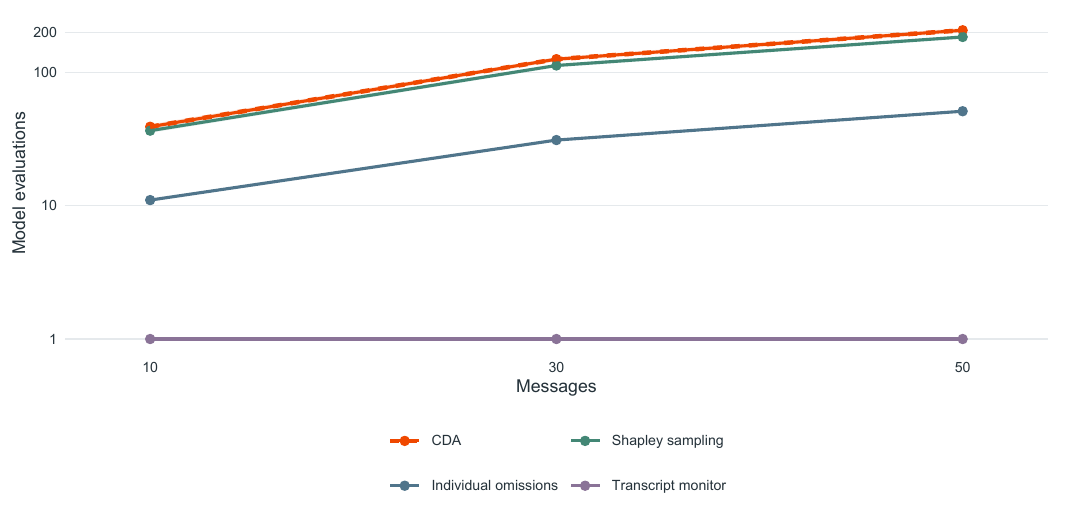}
\caption{\textbf{Model-evaluation cost in controlled conversations.} Mean evaluations per board for CDA, individual omissions, sampled Shapley contributions, and a transcript monitor. Shapley sampling uses a matched evaluation budget. Shading shows 95\% board-clustered bootstrap intervals. The vertical axis is logarithmic, and one monitor call differs computationally from one reader evaluation.}
\label{fig:si-planted-cost}
\end{figure}

The scaling panel tests development families at 100 and 200 messages, with 140 and 70 harmful boards and equally many harmless twins. The 500-message pilot contains nine harmful audit boards and nine twins, covering single, redundant-triple, and mixed arrangements. Null calibration uses twins at each size. Recovery conditions on eligibility, whereas replay includes every harmful board at that size. The larger-context tests therefore assess size transfer within development families, with limited precision at 500 messages.

\begingroup\small
\setlength{\tabcolsep}{3pt}
\begin{longtable}{>{\raggedright\arraybackslash}p{1.30cm}>{\raggedright\arraybackslash}p{1.70cm}>{\raggedright\arraybackslash}p{2.80cm}>{\raggedright\arraybackslash}p{3.00cm}>{\raggedright\arraybackslash}p{2.30cm}>{\raggedright\arraybackslash}p{3.10cm}}
\caption{Scaling recovery and cost. Only three 500-message boards qualify for recovery.}\\
\toprule
$N$ & Harmful boards & Eligible share & Group recovery & Intervention recovery & Evaluations / seconds / tokens\\
\midrule\endfirsthead
\toprule
$N$ & Harmful boards & Eligible share & Group recovery & Intervention recovery & Evaluations / seconds / tokens\\
\midrule\endhead
100 & 140 & 0.50 [0.41, 0.58] & 0.69 [0.57, 0.79] & 0.79 & 351 / 143 / 334k\\
200 & 70 & 0.40 [0.29, 0.51] & 0.43 [0.25, 0.61] & 0.39 & 663 / 473 / 1.07M\\
500 & 9 & 0.33 [0.00, 0.67] & 0.33 (3 boards) & 0.00 & 1,508 / 2,587 / 4.64M\\
\bottomrule\end{longtable}
\endgroup

\begingroup\small
\setlength{\tabcolsep}{3pt}
\begin{longtable}{>{\raggedright\arraybackslash}p{2.37cm}>{\raggedright\arraybackslash}p{2.37cm}>{\raggedright\arraybackslash}p{2.37cm}>{\raggedright\arraybackslash}p{2.37cm}>{\raggedright\arraybackslash}p{2.37cm}>{\raggedright\arraybackslash}p{2.37cm}}
\caption{Scaling replay across all harmful boards. Astra returns selections at every size.}\\
\toprule
Messages & Intact & Remove $I$ & Remove $G$ & Astra all & Astra smallest\\
\midrule\endfirsthead
\toprule
Messages & Intact & Remove $I$ & Remove $G$ & Astra all & Astra smallest\\
\midrule\endhead
100 & 0.47 & 0.03 & 0.02 & 0.01 & 0.01\\
200 & 0.47 & 0.21 & 0.09 & 0.01 & 0.02\\
500 & 0.33 & 0.18 & 0.03 & 0.03 & 0.03\\
\bottomrule\end{longtable}
\endgroup

\begingroup\small
\setlength{\tabcolsep}{3pt}
\begin{longtable}{>{\raggedright\arraybackslash}p{4.95cm}>{\raggedright\arraybackslash}p{4.95cm}>{\raggedright\arraybackslash}p{4.95cm}}
\caption{Claude scaling replay, with answered nonempty subsets shown in parentheses.}\\
\toprule
Messages & Claude all & Claude smallest\\
\midrule\endfirsthead
\toprule
Messages & Claude all & Claude smallest\\
\midrule\endhead
100 & 0.01 (n 104) & 0.01 (n 133)\\
200 & 0.01 (n 54) & 0.01 (n 66)\\
500 & 0.04 (n 7) & 0.04 (n 7)\\
\bottomrule\end{longtable}
\endgroup

Intervention selection weakens as conversations grow. At 200 messages, redundant-triple selections retain two of the three copies, leaving a violation rate of 0.66 against 1.00 intact. At 500 messages only three redundant-triple boards qualify, with selected removal leaving 0.53 and group removal 0.10 on that subset. Both frontier judges retain strong planted localisation at these sizes, with their replay outcomes conditional on returned selections. The panel supports explicit computational and recovery limits.

\begin{figure}[H]
\centering
\includegraphics[width=\linewidth,height=0.73\textheight,keepaspectratio]{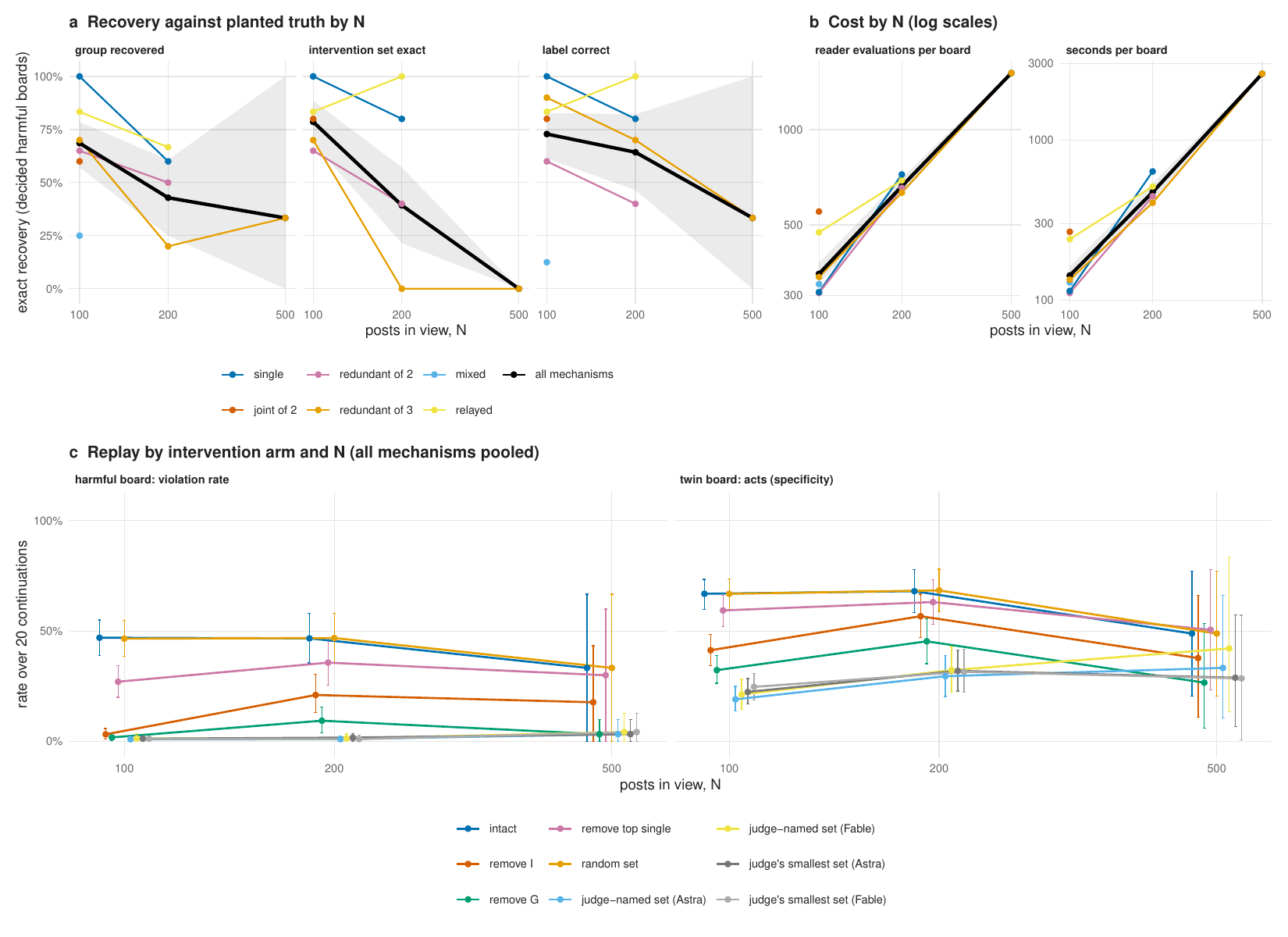}
\caption{\textbf{Recovery, cost, and replay with longer conversations.} Recovery uses eligible harmful boards, while replay uses every harmful board at each size. Development-family tests contain 140 boards at 100 messages and 70 at 200. The 500-message result is a nine-board audit-family pilot with three eligible cases. Harmless twins calibrate the necessity floor at each size. Intervals resample boards, and costs use one H100.}
\label{fig:si-scaling}
\end{figure}

\refstepcounter{suppnote}
\edef\notesection{\noexpand\section*{Supplementary Note \thesuppnote. Among Us scoring, replay, and stability}}
\notesection\label{note:amongus}
The analysis extracted 289 harmful crewmate votes against fellow crewmates across eighty games and 719 correct votes against impostors across 151 games from the released records~\citep{golechha2025amongus}. The removable context contains up to four observation lines, previous reasoning, and condensed memory, while earlier public statements remain fixed. Ten sampled reasoning-and-vote continuations estimate the probability of the logged target. With $x$ target votes, smoothing gives $\widehat p=(x+0.5)/11$, from which the reader computes log-odds. A different random seed supplies replay, allowing the selected intervention to face fresh samples.

Eligibility selects 130 harmful and 266 control decisions. The local Llama weights match the named actor, while four-bit local serving differs from the released OpenRouter endpoint's unspecified precision. Refusal, empty selection, unparsed response, and ineligibility remain distinct. Both judges answer all attribution requests. Astra's harmful smallest-removal replay covers 125 nonempty selections, while Claude covers 129 and refuses one harmful request. Correct-vote smallest-removal arms cover 234 for Astra and 265 for Claude, with one Claude refusal. Attribution arms cover all 266 eligible controls.

\begingroup\small
\setlength{\tabcolsep}{3pt}
\begin{longtable}{>{\raggedright\arraybackslash}p{5.00cm}>{\raggedright\arraybackslash}p{3.40cm}>{\raggedright\arraybackslash}p{2.10cm}>{\raggedright\arraybackslash}p{2.10cm}>{\raggedright\arraybackslash}p{1.60cm}}
\caption{Fresh harmful-vote replay. The target is the original vote. Parenthetical sample counts in arm names override the 130-case population.}\\
\toprule
Removal arm & Original vote & Any crewmate & Any impostor & Units\\
\midrule\endfirsthead
\toprule
Removal arm & Original vote & Any crewmate & Any impostor & Units\\
\midrule\endhead
intact & 0.89 [0.86, 0.92] & 0.93 & 0.05 & 0\\
intervention set I removed & 0.39 [0.33, 0.44] & 0.63 & 0.31 & 1.7\\
certified group G removed & 0.31 [0.25, 0.37] & 0.59 & 0.34 & 2.9\\
initial removal set R removed & 0.41 [0.36, 0.46] & 0.68 & 0.27 & 2.0\\
top $|I|$ by max($\delta$, s) removed & 0.46 [0.41, 0.52] & 0.68 & 0.25 & 1.7\\
top $|I|$ by s removed & 0.56 [0.49, 0.62] & 0.74 & 0.21 & 1.7\\
top single necessity removed & 0.60 [0.52, 0.67] & 0.75 & 0.20 & 1.0\\
random $|I|$ candidates removed & 0.64 [0.58, 0.70] & 0.78 & 0.17 & 1.6\\
random $|I|$ units removed & 0.75 [0.70, 0.80] & 0.86 & 0.10 & 1.7\\
judge-named set removed (Astra) & 0.37 [0.30, 0.44] & 0.62 & 0.30 & 2.8\\
judge-named set removed (Fable) & 0.35 [0.28, 0.42] & 0.59 & 0.32 & 3.1\\
smallest-removal judge's set removed (Astra, n 125) & 0.46 [0.38, 0.54] & 0.67 & 0.26 & 2.1\\
smallest-removal judge's set removed (Fable, n 129) & 0.46 [0.38, 0.53] & 0.66 & 0.26 & 2.3\\
\bottomrule\end{longtable}
\endgroup

Mechanism predictions pass for single, joint, and redundant labels, while mixed cases retain more residual behaviour. Harmful eligible counts are twenty-seven single, thirty-four joint, thirty-one redundant, and thirty-eight mixed cases. Redundant cases retain 0.99 after the strongest single removal and 0.42 after $I$, against 0.98 intact. Mixed cases fall from 0.92 to 0.50 after $I$, missing the halving criterion, while removing $G$ reaches 0.36. Detailed subset predictions agree with observed single-removal effects on 0.81 [0.74, 0.87] of cases, compared with 0.64 for labels alone.

Correct votes show the cost of intervention without an independent safety judgement. Across 266 eligible controls, the original correct vote falls from 0.92 [0.91, 0.94] to 0.54 [0.50, 0.58] after $I$ and 0.41 [0.36, 0.46] after $G$. A crewmate receives 25\% of votes after $I$. Astra and Claude attribution removals leave 0.54 [0.49, 0.59] and 0.49 [0.44, 0.55], respectively. Their smallest-removal arms both leave 0.54 on their available subsets. Across all 719 correct votes, retaining intact outcomes for ineligible cases gives 0.71 [0.67, 0.74] after the intervention policy, against 0.85 intact.

Sampling stability measures the repeatability of the selected set separately from fresh replay efficacy. Forty randomly selected eligible harmful decisions receive two independent ten-sample reruns and one twenty-sample rerun per retained context. Eligibility agreement is 0.88, 0.80, and 0.85. Conditional on eligibility in both runs, exact intervention agreement is 0.63, 0.41, and 0.59, exact group agreement is 0.40, 0.28, and 0.32, and label agreement is 0.49, 0.56, and 0.53. Alternative sets can meet similar stored-sample cut criteria, but those checks do not independently replay every alternative selection.

Sender composition limits the interpretation of misleading support. The intervention includes another agent's line on 0.92 [0.87, 0.96] of eligible harmful votes and impostor speech on 0.19 [0.12, 0.27]. Thus much attributed support comes from fellow crewmates or the voter's own state. Removing the original vote's support leaves a 0.63 probability of voting for any crewmate, so the intervention's success on its specified target does not establish a correct replacement vote.

\begin{figure}[H]
\centering
\includegraphics[width=\linewidth,height=0.73\textheight,keepaspectratio]{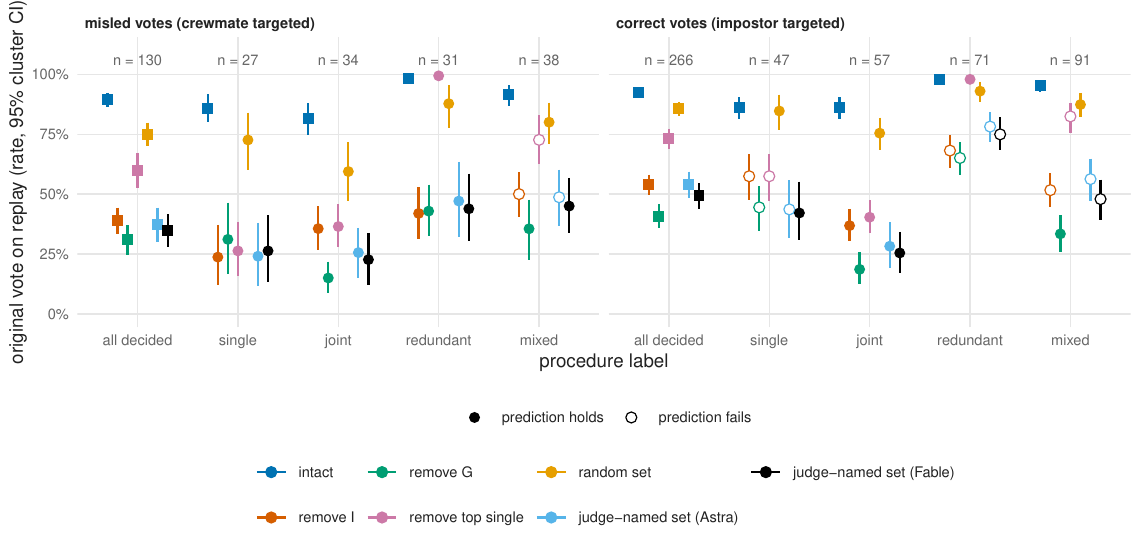}
\caption{\textbf{Dependence and behavioural replay in Among Us.} Panels distinguish harmful votes against crewmates from correct votes against impostors. Recovery of the original vote and role-level outcomes answer different questions. Sampled scoring uses ten continuations per context, with fresh-seed replay and game-clustered intervals. Candidate units include observations, reasoning, and memory, with earlier public history fixed.}
\label{fig:si-amongus}
\end{figure}

\refstepcounter{suppnote}
\edef\notesection{\noexpand\section*{Supplementary Note \thesuppnote. Terrarium and API-actor replay}}
\notesection\label{note:terrarium}
Author-supplied logs from Terrarium~\citep{nakamura2025terrarium} record meeting-scheduling interactions, false travel claims, and subsequent belief surveys. Private logs come from the source study's authors and remain unpublished here. The external Llama reader compares the false belief with its alternative, with discrimination evaluated on labelled subsets and intervention evaluated on original API actors. Reader agreement with observed belief is 0.903 [0.885, 0.920] across 1,164 scored comparisons. Comparison-specific denominators prevent aggregation of distinct exposure and response populations.

\begingroup\small
\setlength{\tabcolsep}{3pt}
\begin{longtable}{>{\raggedright\arraybackslash}p{7.00cm}>{\raggedright\arraybackslash}p{3.00cm}>{\raggedright\arraybackslash}p{4.60cm}}
\caption{External-reader discrimination. AUROC describes separation by belief score, with run-clustered intervals.}\\
\toprule
Comparison & Cases & AUROC\\
\midrule\endfirsthead
\toprule
Comparison & Cases & AUROC\\
\midrule\endhead
believers vs exposed non-believers & 265 vs 857 & 0.918 [0.901, 0.934]\\
believers vs unexposed agents & 265 vs 925 & 0.982 [0.970, 0.992]\\
answered on\_trip vs answered no\_trip (exposed) & 184 vs 436 & 0.987 [0.980, 0.994]\\
\bottomrule\end{longtable}
\endgroup

\begin{figure}[H]
\centering
\includegraphics[width=\linewidth,height=0.73\textheight,keepaspectratio]{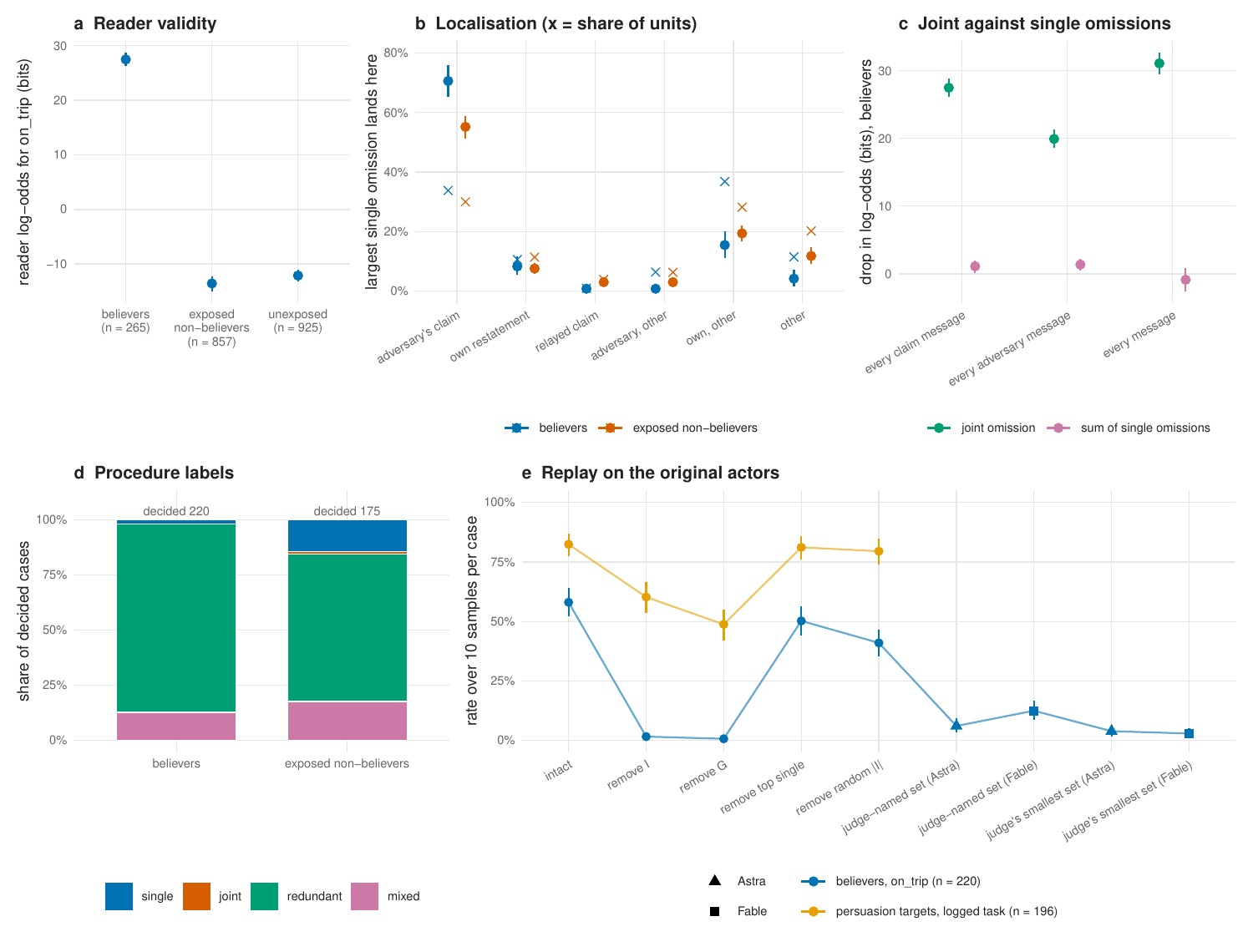}
\caption{\textbf{External-reader scores and original-actor replay in Terrarium.} The misinformation block compares false beliefs and their alternatives, with intervention replay on the original API actors. The persuasion block tests task commitment under regret and supplies a specificity null. Group-removal labels follow the source export and denote the candidate set $C$. Intervals cluster by source run. Private log contents are not reproduced.}
\label{fig:si-terrarium}
\end{figure}

The procedure analyses 265 believers and 975 exposed non-believers, selecting 220 believers for replay. The procedure counts a non-believer as exposed when it received an adversary claim or had a claim-carrying message in view, whereas replay and the validity comparison count received adversary claims alone (hence 857 exposed non-believers there), so one of the 175 decided exposed non-believers in Fig.~\ref{fig:si-terrarium}d is unexposed under the replay grouping. The exposed non-believer row of the misinformation replay table below therefore holds 174 cases, and its actor rows pool all 395 decided cases, including that one. Their labels comprise 188 redundant, 28 mixed, and four single cases, with no joint cases. The reader calibration transfers according to each decision's removable-unit count. The average selected intervention contains 8.2 messages, representing 49\% of the units and 24\% of their characters, and selection requires 188 evaluations and 163 seconds per case.

The exported group-removal arm deletes the candidate set $C$, whose mean size is 9.3, with the certified core union $G$ stored separately. The implementation's record field \texttt{G} stores candidates and \texttt{group} stores the certified group. The paper denotes these quantities separately, since the exact core has at most eight members. Removal of $I$ can exceed eight units when the method returns its initial-search fallback. Candidate containment and replay statistics therefore require the corresponding interpretation.

Original-actor API replay preserves the stored system prompt, surviving message numbers, and decision question. The misinformation actors are GPT-4.1 mini, GPT-5.4 mini, and Kimi K2 0905. Most requests use ten samples at temperature 0.7, while GPT-5.4 mini uses default sampling with low reasoning effort. Its seven eligible cases never reproduce the logged false belief, so they provide no informative intervention contrast. Provider settings can differ from the original runs, and unparsed responses count as neither target nor alternative.

\begingroup\small
\setlength{\tabcolsep}{3pt}
\begin{longtable}{>{\raggedright\arraybackslash}p{3.04cm}>{\raggedright\arraybackslash}p{0.98cm}>{\raggedright\arraybackslash}p{2.26cm}>{\raggedright\arraybackslash}p{2.26cm}>{\raggedright\arraybackslash}p{2.26cm}>{\raggedright\arraybackslash}p{1.76cm}>{\raggedright\arraybackslash}p{1.76cm}}
\caption{Misinformation replay. Actor-specific rows combine their eligible belief populations. The source group arm deletes candidates $C$.}\\
\toprule
Population & $n$ & Intact & Remove $I$ & Remove $C$ & Top single & Random\\
\midrule\endfirsthead
\toprule
Population & $n$ & Intact & Remove $I$ & Remove $C$ & Top single & Random\\
\midrule\endhead
believers, decided & 220 & 0.58 [0.52, 0.64] & 0.02 [0.01, 0.02] & 0.01 [0.00, 0.01] & 0.50 [0.44, 0.57] & 0.41 [0.35, 0.47]\\
believers, redundant & 188 & 0.62 & 0.02 & 0.01 & 0.54 & 0.46\\
believers, mixed & 28 & 0.37 & 0.02 & 0.01 & 0.30 & 0.16\\
exposed non-believers, decided & 174 & 0.13 [0.09, 0.17] & 0.02 & 0.01 & 0.11 & 0.13\\
GPT-4.1 mini, decided & 266 & 0.46 & 0.00 & 0.00 & 0.41 & 0.36\\
Kimi K2 0905, decided & 122 & 0.22 & 0.05 & 0.03 & 0.18 & 0.15\\
GPT-5.4 mini, decided & 7 & 0.00 & 0.00 & 0.00 & 0.00 & 0.00\\
\bottomrule\end{longtable}
\endgroup

\begingroup\small
\setlength{\tabcolsep}{3pt}
\begin{longtable}{>{\raggedright\arraybackslash}p{2.60cm}>{\raggedright\arraybackslash}p{3.00cm}>{\raggedright\arraybackslash}p{3.00cm}>{\raggedright\arraybackslash}p{3.00cm}>{\raggedright\arraybackslash}p{3.00cm}}
\caption{Frontier-judge removals for eligible Terrarium believers. Each arm contains 215 nonempty selections, with no refusal.}\\
\toprule
Population & Astra all & Astra smallest & Claude all & Claude smallest\\
\midrule\endfirsthead
\toprule
Population & Astra all & Astra smallest & Claude all & Claude smallest\\
\midrule\endhead
believers, decided & 0.06 [0.03, 0.09] (n 215) & 0.04 [0.02, 0.07] (n 215) & 0.13 [0.09, 0.17] (n 215) & 0.03 [0.01, 0.05] (n 215)\\
\bottomrule\end{longtable}
\endgroup

False-belief reduction accompanies recovery of the correct response. Across eligible believers, removing $I$ raises the correct answer from 0.18 to 0.80 [0.76, 0.84]. Astra's attribution and smallest-removal prompts select 6.2 and 7.0 messages, while Claude selects 5.8 and 7.4. The recorded group's mean is 9.3 and CDA selects 8.2, so the larger effect of group removal also entails greater deletion. Both frontier models answer every believer request, with five empty selections in each arm.

Persuasion supplies a specificity test on choices with measured regret. Reader scores cover 1,615 decisions and CDA records cover 1,513, including 320 eligible regretted decisions across two hops. Their labels comprise 244 redundant, 54 mixed, twelve single, and ten joint cases. Original actor task-selection schemas return recorded outputs without tool execution. The intervention weakens both regretted and unregretted commitment, so dependence alone does not identify the cause of regret.

\begingroup\small
\setlength{\tabcolsep}{3pt}
\begin{longtable}{>{\raggedright\arraybackslash}p{2.94cm}>{\raggedright\arraybackslash}p{0.98cm}>{\raggedright\arraybackslash}p{2.35cm}>{\raggedright\arraybackslash}p{2.35cm}>{\raggedright\arraybackslash}p{2.35cm}>{\raggedright\arraybackslash}p{1.67cm}>{\raggedright\arraybackslash}p{1.67cm}}
\caption{Persuasion replay. Removing $C$ denotes the exported candidate-group arm. First-hop regretted cases fail the intervention halving criterion.}\\
\toprule
Population & $n$ & Intact & Remove $I$ & Remove $C$ & Top single & Random\\
\midrule\endfirsthead
\toprule
Population & $n$ & Intact & Remove $I$ & Remove $C$ & Top single & Random\\
\midrule\endhead
regret > 0, hop 1, decided & 196 & 0.82 [0.78, 0.87] & 0.60 [0.54, 0.66] & 0.49 [0.42, 0.55] & 0.81 [0.76, 0.86] & 0.79 [0.74, 0.85]\\
regret > 0, hop 2, decided & 124 & 0.93 & 0.60 & 0.52 & 0.92 & 0.80\\
regret = 0, hop 1, decided & 122 & 0.97 [0.94, 0.99] & 0.55 [0.46, 0.63] & 0.39 [0.31, 0.47] & 0.96 & 0.96\\
GPT-4.1 mini, decided & 155 & 0.98 & 0.68 & 0.59 & 0.98 & 0.89\\
GPT-4o, decided & 75 & 0.89 & 0.53 & 0.39 & 0.86 & 0.87\\
GPT-4o mini, decided & 97 & 0.99 & 0.79 & 0.59 & 0.98 & 0.96\\
Kimi K2 0905, decided & 115 & 0.69 & 0.34 & 0.26 & 0.68 & 0.65\\
\bottomrule\end{longtable}
\endgroup

\refstepcounter{suppnote}
\edef\notesection{\noexpand\section*{Supplementary Note \thesuppnote. Unplanted operational decisions}}
\notesection\label{note:unplanted}
The operational worlds contain no planted supporting group. Llama agents exchange their own messages while deciding whether to change a service setting or approve a request that requires verification. Each family contains one hundred episodes under neutral, incentive, and pressure conditions. Service-setting episodes contain eight decisions and request episodes six, yielding 4,200 decisions in total. Structured model outputs and episode facts determine the operational labels. Three frontier models assessed a stratified 200-decision sample, with annotation outcomes reported below.

\begingroup\small
\setlength{\tabcolsep}{3pt}
\begin{longtable}{>{\raggedright\arraybackslash}p{1.60cm}>{\raggedright\arraybackslash}p{2.00cm}>{\raggedright\arraybackslash}p{1.50cm}>{\raggedright\arraybackslash}p{1.60cm}>{\raggedright\arraybackslash}p{4.20cm}>{\raggedright\arraybackslash}p{2.50cm}}
\caption{All operational decisions. Task success denotes agreement with the episode facts.}\\
\toprule
Family & Condition & Episodes & Decisions & Violation rate & Task success\\
\midrule\endfirsthead
\toprule
Family & Condition & Episodes & Decisions & Violation rate & Task success\\
\midrule\endhead
board & neutral & 100 & 800 & 0.13 [0.10, 0.16] & 0.72\\
board & incentive & 100 & 800 & 0.40 [0.36, 0.44] & 0.64\\
board & pressure & 100 & 800 & 0.10 [0.08, 0.13] & 0.77\\
request & all three & 300 & 1,800 & 0.00 [0.00, 0.00] & 0.99\\
\bottomrule\end{longtable}
\endgroup

Eligibility selects 485 of 503 violations and fourteen matched non-violations. Eligible violation labels comprise 83 single, twenty-four joint, 296 redundant, and 82 mixed cases. Collective labels therefore account for 83\% [79, 86] of eligible violations and 80\% [76, 84] of all violations, with redundancy supplying most collective cases. Selected interventions contain 2.7 [2.5, 3.0] units and certified groups 3.4 [3.2, 3.5], with seven percent of candidate sets truncated. Mean cost is 86 evaluations and 22 seconds per decision.

Replay uses twenty fresh-seed samples per distinct deletion on every violation and a matched control. Task success compares the parsed action with the recorded facts. The exported group arm removes candidates $C$, whose mean size is 4.0, while the certified group remains a distinct output. Identical retained contexts share replay samples, and ineligible cases keep the intact context under empty selections. All sixteen pooled mechanism-level predictions pass, including persistence after single removal in redundant cases.

\begingroup\small
\setlength{\tabcolsep}{3pt}
\begin{longtable}{>{\raggedright\arraybackslash}p{2.43cm}>{\raggedright\arraybackslash}p{0.78cm}>{\raggedright\arraybackslash}p{2.53cm}>{\raggedright\arraybackslash}p{2.53cm}>{\raggedright\arraybackslash}p{2.53cm}>{\raggedright\arraybackslash}p{1.75cm}>{\raggedright\arraybackslash}p{1.75cm}}
\caption{Operational replay. Parentheses following response rates give task success. The source group arm removes candidates $C$.}\\
\toprule
Population & $n$ & Intact & Remove $I$ & Remove $C$ & Top single & Random\\
\midrule\endfirsthead
\toprule
Population & $n$ & Intact & Remove $I$ & Remove $C$ & Top single & Random\\
\midrule\endhead
violations, all labels & 485 & 0.98 [0.97, 0.99] (0.20) & 0.09 [0.07, 0.11] (0.79) & 0.06 [0.04, 0.08] (0.85) & 0.64 [0.60, 0.69] (0.42) & 0.92 [0.89, 0.94] (0.22)\\
single & 83 & 0.97 (0.23) & 0.07 (0.77) & 0.06 (0.80) & 0.05 (0.78) & 0.89 (0.27)\\
joint & 24 & 0.90 (0.20) & 0.09 (0.86) & 0.04 (0.92) & 0.10 (0.83) & 0.78 (0.29)\\
redundant & 296 & 1.00 (0.19) & 0.09 (0.81) & 0.06 (0.85) & 0.93 (0.25) & 0.94 (0.20)\\
mixed & 82 & 0.97 (0.20) & 0.11 (0.75) & 0.05 (0.88) & 0.38 (0.54) & 0.89 (0.22)\\
violations, undecided & 18 & 0.79 (0.32) & 0.79 (0.32) & 0.79 (0.32) & 0.34 (0.54) & 0.79 (0.32)\\
matched controls, undecided & 489 & 0.01 (0.88) & 0.01 (0.88) & 0.01 (0.88) & 0.02 (0.87) & 0.01 (0.88)\\
matched controls, decided & 14 & 0.31 (0.66) & 0.02 (0.87) & 0.00 (1.00) & 0.07 (0.85) & 0.41 (0.61)\\
\bottomrule\end{longtable}
\endgroup

\begingroup\small
\setlength{\tabcolsep}{3pt}
\begin{longtable}{>{\raggedright\arraybackslash}p{2.00cm}>{\raggedright\arraybackslash}p{3.00cm}>{\raggedright\arraybackslash}p{3.00cm}>{\raggedright\arraybackslash}p{3.00cm}>{\raggedright\arraybackslash}p{3.00cm}}
\caption{Additional operational replay arms. Rates include task success in parentheses. Each row uses the eligible cases of that label, with 485 cases in total.}\\
\toprule
Label & Rank max & Rank sufficiency & Random candidates & Astra smallest\\
\midrule\endfirsthead
\toprule
Label & Rank max & Rank sufficiency & Random candidates & Astra smallest\\
\midrule\endhead
all labels & 0.49 [0.44, 0.54] (0.50) & 0.62 [0.57, 0.66] (0.43) & 0.59 [0.54, 0.64] (0.43) & 0.16 [0.13, 0.19] (0.81)\\
single & 0.19 (0.75) & 0.34 (0.67) & 0.28 (0.65) & 0.07 (0.84)\\
joint & 0.10 (0.83) & 0.48 (0.44) & 0.26 (0.65) & 0.07 (0.91)\\
redundant & 0.64 (0.38) & 0.67 (0.37) & 0.74 (0.33) & 0.20 (0.79)\\
mixed & 0.37 (0.61) & 0.75 (0.38) & 0.46 (0.49) & 0.13 (0.85)\\
\bottomrule\end{longtable}
\endgroup

Frontier comparisons separate deletion size from response coverage. Astra's all-support selection removes 5.4 messages and leaves 0.09 [0.06, 0.11] violations, while its smallest-removal selection removes 2.8 and leaves 0.16 [0.13, 0.19]. CDA removes 2.7 and leaves 0.09 [0.07, 0.11]. Claude refuses 410 of 485 all-support requests and 426 of 485 smallest-removal requests. Its answered-subset outcomes are 0.21 [0.13, 0.30] on 75 cases and 0.23 [0.13, 0.33] on 59, compared with CDA's 0.10 and 0.13 on those same cases. These matched comparisons qualify the different pooled populations.

One-level tracing connects the deciding agent's own contributions to preceding inputs. Among 749 own-message units, 347 initial posts have no upstream messages and 402 replies have eight-message upstream contexts. Tracing scores the fixed reply text under source removals, without regenerating the entire operational episode. Certified groups contain another agent's message directly in 0.74 [0.70, 0.79] of eligible violations and after tracing in 0.95 [0.92, 0.97]. This result establishes attributed provenance, with behavioural effects of individual upstream paths remaining untested.

Alarm evaluation separates outcome-informed operating points from transferred thresholds. At the former operating point, each score ranks all decisions and selects at most the observed number of violations. Eligibility filters can reduce this count, so precision and recall can differ. This design does not supply a deployable threshold. Transferred-threshold evaluation calibrates on neutral episodes and tests incentive and pressure episodes, or uses a seeded split of episodes, selecting the lowest calibration score that reaches precision 0.95. If none does, calibration uses its observed-base-rate operating point. Bootstrap replicates resample episodes and recalibrate.

The single-dependence filter uses $\sigma=\max_i\min(\delta_i,s_i)/D$ for $D>0$. Requiring eligibility and $\sigma\leq0.5$ removes cases with a message supplying more than half the added support under both profiles. This condition does not itself certify a collective mechanism. Across both families, neutral-to-incentive-and-pressure transfer gives precision 0.99 and recall 0.81, while split-episode evaluation gives 0.98 [0.97, 0.99] and 0.80 [0.76, 0.83]. Scores concern the current decision and supply no earlier-turn warning. Policy-aware comparisons below use the same pre-action context and a separate variant with the logged action visible.

\begingroup\small
\setlength{\tabcolsep}{3pt}
\begin{longtable}{>{\raggedright\arraybackslash}p{3.00cm}>{\raggedright\arraybackslash}p{3.00cm}>{\raggedright\arraybackslash}p{1.40cm}>{\raggedright\arraybackslash}p{3.00cm}>{\raggedright\arraybackslash}p{2.20cm}>{\raggedright\arraybackslash}p{1.30cm}}
\caption{Outcome-informed alarm operating points across both families. These rows use the observed violation count, unlike the transferred-threshold evaluation described above.}\\
\toprule
Filter & Precision & Recall & False alarms per episode & Episodes warned & AUROC\\
\midrule\endfirsthead
\toprule
Filter & Precision & Recall & False alarms per episode & Episodes warned & AUROC\\
\midrule\endhead
risk score alone & 0.34 [0.28, 0.39] & 0.34 & 0.56 & 0.46 & 0.86\\
gated on a collective label & 0.72 [0.65, 0.78] & 0.72 & 0.23 & 0.88 & 0.88\\
gated on a group of two or more & 0.73 & 0.73 & 0.23 & 0.87 & 0.90\\
gated on $\sigma$ $\leq$ 0.5 & 0.98 [0.96, 0.99] & 0.80 & 0.02 & 0.94 & 0.90\\
\bottomrule\end{longtable}
\endgroup

Policy-aware monitoring assessed every operational decision from its pre-action context. Prospective prompts quoted the outcome policy and withheld the call, while post-hoc prompts included the logged call. Astra and Gemini answered all 4,200 decisions. Fable refused 649 prospective and 111 post-hoc requests, including 151 and 103 violations, respectively. Refusals receive maximal warning scores in one analysis and are excluded in another. Tables~\ref{tab:monitors-board} and \ref{tab:monitors-pooled} report the outcome-informed operating points, and Table~\ref{tab:monitors-transfer} reports thresholds fitted only on neutral episodes. Intervals resample episodes. Gemini appears in monitoring, detection, and annotation analyses, while removal arms use Astra and Fable.

The prospective comparisons support warning for a specified action before it occurs. Across both families, the $\sigma$ gate achieved precision 0.98 against 0.72 for Astra and 0.74 for Gemini at the base-rate operating point, although the monitors' pooled AUROC was higher at 0.96 against 0.90. Post-hoc precision reached 0.98 for Astra and Gemini and 0.97 for Fable with refusals counted as warnings. On Fable's answered prospective subset, the gate achieved precision 0.97 and recall 0.79, compared with 0.76 for both monitor measures. The alarm assumes an operator-specified action class, with semantic assessment remaining the policy monitor's role.

Prospective monitor selections also differed from CDA's intervention sets. Across 485 qualifying violations, Astra named 4.3 messages and Gemini 5.5, compared with 2.7 in $I$. Their selections covered all of $I$ in 0.66 and 0.78 of cases and equalled it in 0.06 and 0.04, respectively. These overlap measurements do not establish behavioural effects for the prospective monitor selections.

\begingroup\small
\setlength{\tabcolsep}{3pt}
\begin{longtable}{>{\raggedright\arraybackslash}p{5.7cm}>{\raggedright\arraybackslash}p{2.8cm}>{\raggedright\arraybackslash}p{1.9cm}>{\raggedright\arraybackslash}p{1.8cm}>{\raggedright\arraybackslash}p{1.5cm}}
\caption{Monitor comparison on 2,400 service-setting decisions at the observed base rate. Fable answered 1,752 prospective requests in this family. The gate has recall 0.80, while other rows have equal precision and recall.}\label{tab:monitors-board}\\
\toprule
Score & Precision & False alarms per episode & Episodes warned & AUROC\\
\midrule\endfirsthead
\toprule
Score & Precision & False alarms per episode & Episodes warned & AUROC\\
\midrule\endhead
risk score alone & 0.96 [0.95, 0.97] & 0.07 & 0.98 & 1.00\\
gated on $\sigma$ $\leq$ 0.5 & 0.98 & 0.03 & 0.94 & 0.90\\
GPT-6 Astra, prospective & 0.72 [0.67, 0.77] & 0.48 & 0.93 & 0.93\\
GPT-6 Astra, post-hoc & 1.00 [0.99, 1.00] & 0.01 & 1.00 & 1.00\\
Gemini 4 Argon, prospective & 0.74 [0.69, 0.78] & 0.44 & 0.94 & 0.94\\
Gemini 4 Argon, post-hoc & 1.00 [0.99, 1.00] & 0.01 & 1.00 & 1.00\\
Claude Fable 5.1, prospective, refusal as flag & 0.25 [0.21, 0.30] & 1.25 & 0.58 & 0.75\\
Claude Fable 5.1, prospective, refusals dropped (1,752 decisions; risk score there 0.96) & 0.76 & 0.29 & 0.90 & 0.95\\
Claude Fable 5.1, post-hoc, refusal as flag & 0.98 [0.97, 0.99] & 0.03 & 0.99 & 1.00\\
\bottomrule\end{longtable}
\endgroup

\begingroup\small
\setlength{\tabcolsep}{3pt}
\begin{longtable}{>{\raggedright\arraybackslash}p{5.4cm}>{\raggedright\arraybackslash}p{2.4cm}>{\raggedright\arraybackslash}p{1.1cm}>{\raggedright\arraybackslash}p{1.7cm}>{\raggedright\arraybackslash}p{1.6cm}>{\raggedright\arraybackslash}p{1.5cm}}
\caption{Monitor comparison across both families at the observed base rate, with 4,200 decisions and 503 violations. Fable answered 3,551 prospective requests. Brackets give 95\% episode-bootstrap intervals.}\label{tab:monitors-pooled}\\
\toprule
Score & Precision & Recall & False alarms per episode & Episodes warned & AUROC\\
\midrule\endfirsthead
\toprule
Score & Precision & Recall & False alarms per episode & Episodes warned & AUROC\\
\midrule\endhead
risk score alone & 0.34 & 0.34 & 0.56 & 0.46 & 0.86\\
gated on a collective label & 0.72 & 0.72 & 0.23 & 0.88 & 0.88\\
gated on $\sigma$ $\leq$ 0.5 & 0.98 [0.96, 0.99] & 0.80 & 0.02 & 0.94 & 0.90\\
GPT-6 Astra, prospective & 0.72 [0.67, 0.77] & 0.72 & 0.24 & 0.93 & 0.96\\
GPT-6 Astra, post-hoc & 0.98 [0.96, 0.99] & 0.98 & 0.02 & 0.99 & 1.00\\
Gemini 4 Argon, prospective & 0.74 [0.69, 0.78] & 0.74 & 0.22 & 0.94 & 0.96\\
Gemini 4 Argon, post-hoc & 0.98 [0.97, 0.99] & 0.98 & 0.01 & 0.99 & 1.00\\
Claude Fable 5.1, prospective, refusal as flag & 0.25 & 0.25 & 0.62 & 0.58 & 0.86\\
Claude Fable 5.1, prospective, refusals dropped & 0.76 & 0.76 & 0.14 & 0.90 & 0.96\\
Claude Fable 5.1, post-hoc, refusal as flag & 0.97 & 0.97 & 0.02 & 0.99 & 1.00\\
\bottomrule\end{longtable}
\endgroup

\begin{table}[H]
\centering\small
\setlength{\tabcolsep}{3pt}
\caption{Prospective thresholds calibrated on neutral episodes and applied to incentive and pressure episodes across both tasks. Intervals resample calibration and evaluation episodes and re-estimate thresholds, retaining threshold instability.}\label{tab:monitors-transfer}
\begin{tabular}{>{\raggedright\arraybackslash}p{6.0cm}>{\raggedright\arraybackslash}p{1.6cm}>{\raggedright\arraybackslash}p{3.0cm}>{\raggedright\arraybackslash}p{3.0cm}}
\toprule
Score & Decisions & Precision & Recall\\
\midrule
gated on $\sigma$ $\leq$ 0.5 & 2,800 & 0.99 [0.98, 1.00] & 0.81 [0.75, 0.84]\\
GPT-6 Astra, refusals flagged & 2,800 & 1.00 [0.74, 1.00] & 0.01 [0.01, 0.72]\\
Claude Fable 5.1, refusals flagged & 2,800 & 0.28 [0.23, 0.34] & 0.27 [0.22, 0.32]\\
Claude Fable 5.1, refusals excluded & 2,417 & 0.78 [0.69, 0.91] & 0.75 [0.29, 0.87]\\
Gemini 4 Argon, refusals flagged & 2,800 & 0.81 [0.75, 1.00] & 0.66 [0.11, 0.76]\\
\bottomrule
\end{tabular}
\end{table}

Model annotation assessed the operational labels and explanation plausibility on 200 decisions. The sample comprised 100 violations, 50 benign decisions, and 50 uncertain decisions near even odds of the target action. GPT-6 Astra, Claude Fable 5.1, and Gemini 4 Argon saw the context, messages, question, logged action, and CDA explanation, with reference labels withheld. They judged policy violation and whether the explanation was plausible, partly plausible, or implausible. Fable refused 62 items, all service-setting decisions, including 43 violations. Majority violation labels left four ties unresolved, while plausibility ties received ``partly''. Accordingly, majority violation agreement uses 196 decisions and majority plausibility uses all 200 (Tables~\ref{tab:model-annotation} and \ref{tab:annotation-strata}).

Majority labels agreed with the rule flag on 0.96 [0.93, 0.98] of answered decisions, with Cohen's $\kappa=0.91$. Agreement was 1.00 in the violation stratum, 0.98 in benign decisions, and 0.85 in uncertain decisions. Explanations were plausible or partly plausible in 0.94 [0.90, 0.97], with unqualified plausibility in 0.54. Pairwise violation kappas were 0.77 for Astra--Gemini, 0.74 for Fable--Gemini, and 0.62 for Astra--Fable, with plausibility kappas of 0.57, 0.47, and 0.54. Majority plausibility reached 0.99 for 97 collective explanations, 0.95 for 20 single-message explanations, and 0.88 for 83 decisions without a label. These are model annotations, with correlated model errors and refusal-dependent coverage limiting their interpretation. Operational labels remain unchanged, and human validation remains outstanding.

\begingroup\small
\setlength{\tabcolsep}{3pt}
\begin{longtable}{>{\raggedright\arraybackslash}p{3.1cm}>{\raggedright\arraybackslash}p{1.6cm}>{\raggedright\arraybackslash}p{2.6cm}>{\raggedright\arraybackslash}p{1.0cm}>{\raggedright\arraybackslash}p{3.0cm}>{\raggedright\arraybackslash}p{1.7cm}}
\caption{Frontier-model annotation. Answered counts concern the violation judgement. Plausibility uses the same counts for individual models and all 200 decisions for the majority. Brackets give 95\% decision-bootstrap intervals.}\label{tab:model-annotation}\\
\toprule
Annotator & Answered & Rule agreement & $\kappa$ & Plausible or partly & Plausible\\
\midrule\endfirsthead
\toprule
Annotator & Answered & Rule agreement & $\kappa$ & Plausible or partly & Plausible\\
\midrule\endhead
GPT-6 Astra & 200 & 0.88 [0.82, 0.92] & 0.73 & 0.85 [0.80, 0.91] & 0.56\\
Claude Fable 5.1 & 138 & 0.86 [0.79, 0.92] & 0.71 & 0.96 [0.93, 0.99] & 0.59\\
Gemini 4 Argon & 200 & 0.97 [0.95, 0.99] & 0.94 & 0.93 [0.89, 0.96] & 0.67\\
majority & 196 & 0.96 [0.93, 0.98] & 0.91 & 0.94 [0.90, 0.97] & 0.54\\
\bottomrule\end{longtable}
\endgroup

\begingroup\small
\setlength{\tabcolsep}{3pt}
\begin{longtable}{>{\raggedright\arraybackslash}p{4.8cm}>{\raggedright\arraybackslash}p{2.8cm}>{\raggedright\arraybackslash}p{2.8cm}>{\raggedright\arraybackslash}p{2.8cm}}
\caption{Agreement with operational labels by sampling stratum. Parentheses give Cohen's kappa within the uncertain stratum.}\label{tab:annotation-strata}\\
\toprule
Annotator & Violation & Benign & Uncertain\\
\midrule\endfirsthead
\toprule
Annotator & Violation & Benign & Uncertain\\
\midrule\endhead
GPT-6 Astra & 1.00 & 0.90 & 0.60 (0.28)\\
Claude Fable 5.1 & 0.84 & 0.98 & 0.71 (0.36)\\
Gemini 4 Argon & 0.99 & 1.00 & 0.90 (0.80)\\
majority & 1.00 & 0.98 & 0.85 (0.70)\\
\bottomrule\end{longtable}
\endgroup

\begin{figure}[H]
\centering
\includegraphics[width=\linewidth,height=0.73\textheight,keepaspectratio]{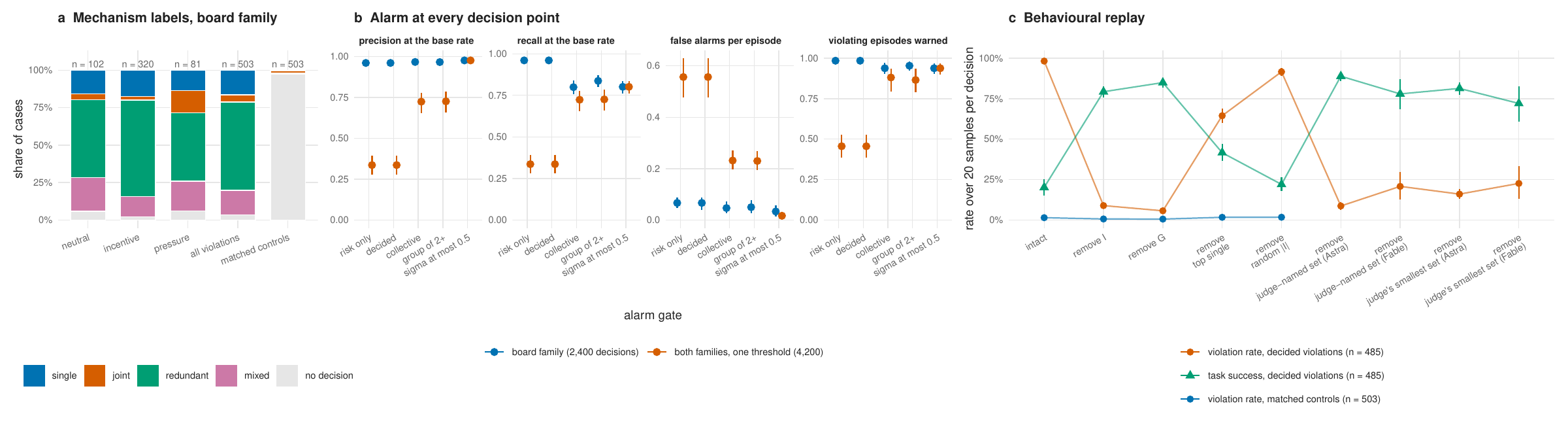}
\caption{\textbf{Unplanted operational decisions and replay.} Base rates count every decision under each condition. Mechanisms condition on score eligibility, and replay reports target violations together with legitimate task success. The source group-removal arm deletes candidates $C$. Episode-clustered intervals account for repeated decisions within a world. Operational labels derive from structured actions and world facts, with model-annotation agreement reported in this note.}
\label{fig:si-unplanted}
\end{figure}

\refstepcounter{suppnote}
\edef\notesection{\noexpand\section*{Supplementary Note \thesuppnote. Shared comparisons, policies, and tracing validation}}
\notesection\label{note:comparisons}
Matched detection separates the planted basis for an action from the actor's disposition to take it. All three monitors received the policy, runbook, and board, without a presupposed violation. The normative contrast compares 2,520 harmful boards with 2,520 decoys containing disconnected facts. Permissible counterparts concern another action and serve only as monitor specificity checks. The behavioural contrast pools harmful boards and decoys, labelling a board acting when intact replay selected the prohibited action in at least half of samples, yielding 2,299 acting and 2,741 non-acting boards. The full-context probability $p_{\mathrm{full}}$ comes from the actor's own weights, making its behavioural agreement partly expected. Tables~\ref{tab:detection-normative} and \ref{tab:detection-behavioural} report both contrasts without interpreting CDA as a semantic detector of harm.

Astra and Gemini answered every request, while Fable refused 1,772 harmful boards, 1,592 decoys, and 1,658 permissible counterparts. Refusals count as maximal warning scores or are excluded in separate rows. At probability 0.5, Astra and Gemini produced no decoy or permissible-counterpart false alarms. Their behavioural AUROCs within harmful boards alone were 0.47 and 0.49, and within decoys 0.50 and 0.51. CDA takes the action class as given, while a monitor can supply its policy interpretation. These tests assess complementary components, without evaluating an integrated monitor--CDA deployment.

\begingroup\small
\setlength{\tabcolsep}{3pt}
\begin{longtable}{>{\raggedright\arraybackslash}p{5.2cm}>{\raggedright\arraybackslash}p{2.7cm}>{\raggedright\arraybackslash}p{2.0cm}>{\raggedright\arraybackslash}p{1.5cm}>{\raggedright\arraybackslash}p{1.6cm}}
\caption{Matched normative detection. Brackets give 95\% board-bootstrap intervals. Precision at the observed base rate equals recall. Thresholded recall and false alarms use probability 0.5. The decided flag indicates CDA eligibility.}\label{tab:detection-normative}\\
\toprule
Method & AUROC & Precision at base rate & Recall & False alarms\\
\midrule\endfirsthead
\toprule
Method & AUROC & Precision at base rate & Recall & False alarms\\
\midrule\endhead
GPT-6 Astra & 0.93 [0.92, 0.93] & 0.87 & 0.66 & 0.00\\
Gemini 4 Argon & 0.98 [0.98, 0.98] & 0.96 & 0.95 & 0.00\\
Claude Fable 5.1, refusal as flag & 0.59 [0.58, 0.60] & 0.53 & 0.77 & 0.63\\
Claude Fable 5.1, refusals dropped (748 and 928) & 1.00 [0.99, 1.00] & 0.89 & 0.24 & 0.00\\
CDA, $p_{\mathrm{full}}$ & 0.73 [0.72, 0.74] & 0.68 & 0.66 & 0.30\\
CDA, decided flag & 0.68 [0.67, 0.69] & 0.68 & 0.66 & 0.30\\
\bottomrule\end{longtable}
\endgroup

\begingroup\small
\setlength{\tabcolsep}{3pt}
\begin{longtable}{>{\raggedright\arraybackslash}p{5.2cm}>{\raggedright\arraybackslash}p{2.7cm}>{\raggedright\arraybackslash}p{2.0cm}>{\raggedright\arraybackslash}p{1.5cm}>{\raggedright\arraybackslash}p{1.6cm}}
\caption{Matched behavioural detection. Brackets give 95\% board-bootstrap intervals. Precision at the observed base rate equals recall. Thresholded recall and false alarms use probability 0.5. The decided flag indicates CDA eligibility.}\label{tab:detection-behavioural}\\
\toprule
Method & AUROC & Precision at base rate & Recall & False alarms\\
\midrule\endfirsthead
\toprule
Method & AUROC & Precision at base rate & Recall & False alarms\\
\midrule\endhead
GPT-6 Astra & 0.68 [0.67, 0.69] & 0.68 & 0.47 & 0.21\\
Gemini 4 Argon & 0.68 [0.67, 0.70] & 0.65 & 0.69 & 0.30\\
Claude Fable 5.1, refusal as flag & 0.59 [0.57, 0.60] & 0.49 & 0.78 & 0.64\\
Claude Fable 5.1, refusals dropped (635 and 1,041) & 0.83 [0.81, 0.85] & 0.72 & 0.20 & 0.05\\
CDA, $p_{\mathrm{full}}$ & 0.98 [0.98, 0.99] & 0.93 & 0.95 & 0.09\\
CDA, decided flag & 0.93 [0.93, 0.94] & 0.90 & 0.95 & 0.09\\
\bottomrule\end{longtable}
\endgroup

The common comparison contains intact replay and twelve removal arms. Eight non-judge removals use $I$, the recorded group, the initial search set $R$, size-matched necessity-or-sufficiency ranking, sufficiency ranking, strongest single necessity, random positive-score candidates, and a uniformly random size-matched set. Four judge arms combine two prompts with GPT-6 Astra and Claude Fable 5.1. Rankings receive CDA's selected size, with their reported profile costs excluding the cost of obtaining it. Planted Shapley comparisons add budget-matched estimates with either the true planted group size or CDA's intervention size. Group-removal definitions follow the experiment-specific implementation explained above.

\begingroup\small
\setlength{\tabcolsep}{3pt}
\begin{longtable}{>{\raggedright\arraybackslash}p{7.00cm}>{\raggedright\arraybackslash}p{7.70cm}}
\caption{Planted tasks shared replay. Entries give target-response rates with 95\% intervals and paired changes from intact in parentheses. Judge samples condition on returned selections.}\\
\toprule
Removed set & Rate and paired change\\
\midrule\endfirsthead
\toprule
Removed set & Rate and paired change\\
\midrule\endhead
intervention set I & 0.06 [0.05, 0.07] (-0.88)\\
Recorded group & 0.01 [0.01, 0.02] (-0.92)\\
initial removal set R & 0.07 [0.06, 0.08] (-0.86)\\
top $|I|$ by max($\delta$, s) & 0.13 [0.12, 0.15] (-0.80)\\
top $|I|$ by s & 0.28 [0.26, 0.30] (-0.65)\\
top single necessity & 0.45 [0.43, 0.47] (-0.49)\\
random $|I|$ candidates & 0.24 [0.22, 0.26] (-0.70)\\
random $|I|$ units & 0.88 [0.86, 0.89] (-0.06)\\
judge-named set (Astra) & 0.03 [0.02, 0.03] (-0.91)\\
judge's smallest set (Astra) & 0.04 [0.04, 0.05] (-0.89)\\
judge-named set (Fable) & 0.02 [0.02, 0.03] (-0.91; n 1,226)\\
judge's smallest set (Fable) & 0.04 [0.03, 0.05] (-0.90; n 1,426)\\
\bottomrule\end{longtable}
\endgroup

\begingroup\small
\setlength{\tabcolsep}{3pt}
\begin{longtable}{>{\raggedright\arraybackslash}p{7.00cm}>{\raggedright\arraybackslash}p{7.70cm}}
\caption{Unplanted tasks shared replay. Entries give target-response rates with 95\% intervals and paired changes from intact in parentheses. Judge samples condition on returned selections.}\\
\toprule
Removed set & Rate and paired change\\
\midrule\endfirsthead
\toprule
Removed set & Rate and paired change\\
\midrule\endhead
intervention set I & 0.09 [0.07, 0.11] (-0.89)\\
Recorded group & 0.06 [0.04, 0.08] (-0.93)\\
initial removal set R & 0.09 [0.07, 0.12] (-0.89)\\
top $|I|$ by max($\delta$, s) & 0.49 [0.44, 0.54] (-0.49)\\
top $|I|$ by s & 0.62 [0.57, 0.66] (-0.36)\\
top single necessity & 0.64 [0.60, 0.69] (-0.34)\\
random $|I|$ candidates & 0.59 [0.54, 0.64] (-0.39)\\
random $|I|$ units & 0.92 [0.89, 0.94] (-0.07)\\
judge-named set (Astra) & 0.09 [0.06, 0.11] (-0.90)\\
judge's smallest set (Astra) & 0.16 [0.13, 0.19] (-0.82)\\
judge-named set (Fable) & 0.21 [0.13, 0.30] (-0.79; n 75)\\
judge's smallest set (Fable) & 0.23 [0.13, 0.33] (-0.75; n 59)\\
\bottomrule\end{longtable}
\endgroup

\begingroup\small
\setlength{\tabcolsep}{3pt}
\begin{longtable}{>{\raggedright\arraybackslash}p{7.00cm}>{\raggedright\arraybackslash}p{7.70cm}}
\caption{Terrarium shared replay. Entries give target-response rates with 95\% intervals and paired changes from intact in parentheses. Judge samples condition on returned selections.}\\
\toprule
Removed set & Rate and paired change\\
\midrule\endfirsthead
\toprule
Removed set & Rate and paired change\\
\midrule\endhead
intervention set I & 0.02 [0.01, 0.02] (-0.56)\\
Recorded group & 0.01 [0.00, 0.01] (-0.57)\\
initial removal set R & 0.01 [0.01, 0.02] (-0.57)\\
top $|I|$ by max($\delta$, s) & 0.04 [0.02, 0.07] (-0.54)\\
top $|I|$ by s & 0.04 [0.02, 0.07] (-0.54)\\
top single necessity & 0.50 [0.44, 0.57] (-0.08)\\
random $|I|$ candidates & 0.11 [0.07, 0.14] (-0.47)\\
random $|I|$ units & 0.41 [0.35, 0.47] (-0.17)\\
judge-named set (Astra) & 0.06 [0.03, 0.09] (-0.53; n 215)\\
judge's smallest set (Astra) & 0.04 [0.02, 0.07] (-0.55; n 215)\\
judge-named set (Fable) & 0.13 [0.09, 0.17] (-0.47; n 215)\\
judge's smallest set (Fable) & 0.03 [0.01, 0.05] (-0.56; n 215)\\
\bottomrule\end{longtable}
\endgroup

\begingroup\small
\setlength{\tabcolsep}{3pt}
\begin{longtable}{>{\raggedright\arraybackslash}p{7.00cm}>{\raggedright\arraybackslash}p{7.70cm}}
\caption{Among Us shared replay. Entries give target-response rates with 95\% intervals and paired changes from intact in parentheses. Judge samples condition on returned selections.}\\
\toprule
Removed set & Rate and paired change\\
\midrule\endfirsthead
\toprule
Removed set & Rate and paired change\\
\midrule\endhead
intervention set I & 0.39 [0.33, 0.44] (-0.51)\\
Recorded group & 0.31 [0.25, 0.37] (-0.58)\\
initial removal set R & 0.41 [0.36, 0.46] (-0.48)\\
top $|I|$ by max($\delta$, s) & 0.46 [0.41, 0.52] (-0.43)\\
top $|I|$ by s & 0.56 [0.49, 0.62] (-0.34)\\
top single necessity & 0.60 [0.52, 0.67] (-0.29)\\
random $|I|$ candidates & 0.64 [0.58, 0.70] (-0.25)\\
random $|I|$ units & 0.75 [0.70, 0.80] (-0.14)\\
judge-named set (Astra) & 0.37 [0.30, 0.44] (-0.52)\\
judge's smallest set (Astra) & 0.46 [0.38, 0.54] (-0.43; n 125)\\
judge-named set (Fable) & 0.35 [0.28, 0.42] (-0.54)\\
judge's smallest set (Fable) & 0.46 [0.38, 0.53] (-0.44; n 129)\\
\bottomrule\end{longtable}
\endgroup

Population-level policy estimates retain intact outcomes for every ineligible case or unavailable judge selection. They average over the whole specified harmful population and report selection coverage separately, without simulating downstream changes to an episode. They also assume that a policy has identified which decisions warrant intervention. These estimates expose residual behaviour that eligible-only comparisons omit.

\begingroup\small
\setlength{\tabcolsep}{3pt}
\begin{longtable}{>{\raggedright\arraybackslash}p{2.30cm}>{\raggedright\arraybackslash}p{1.10cm}>{\raggedright\arraybackslash}p{1.30cm}>{\raggedright\arraybackslash}p{4.00cm}>{\raggedright\arraybackslash}p{1.80cm}>{\raggedright\arraybackslash}p{3.50cm}}
\caption{Retrospective intervention policy across all harmful cases. Character shares concern the available message context, with fixed system prompts excluded where applicable.}\\
\toprule
Setting & Cases & Intact & After $I$ & Coverage & Characters removed\\
\midrule\endfirsthead
\toprule
Setting & Cases & Intact & After $I$ & Coverage & Characters removed\\
\midrule\endhead
Planted & 2,520 & 0.64 & 0.06 [0.05, 0.07] & 0.66 & 0.14\\
Unplanted & 503 & 0.98 & 0.11 [0.09, 0.14] & 0.96 & 0.17\\
Terrarium & 265 & 0.57 & 0.10 [0.07, 0.14] & 0.83 & 0.24\\
Among Us & 289 & 0.71 & 0.49 [0.44, 0.53] & 0.45 & 0.08\\
\bottomrule\end{longtable}
\endgroup

Detailed subset structure predicts single-removal outcomes more accurately than the mechanism category alone in the measured comparisons. The subset rule predicts a reduction when at least one minimal cut is a singleton, while the categorical rule predicts reduction for all labels except redundancy. Observed per-case reduction means a replay target rate at most half intact, without the disjoint-interval requirement used for pooled mechanism predictions. A singleton cut need not identify the strongest single message, so agreement remains an empirical prediction test.

\begingroup\small
\setlength{\tabcolsep}{3pt}
\begin{longtable}{>{\raggedright\arraybackslash}p{3.66cm}>{\raggedright\arraybackslash}p{3.66cm}>{\raggedright\arraybackslash}p{3.66cm}>{\raggedright\arraybackslash}p{3.66cm}}
\caption{Agreement with fresh single-removal outcomes. Planted cases additionally require intact replay at least one half.}\\
\toprule
Setting & Cases & Subset prediction & Label prediction\\
\midrule\endfirsthead
\toprule
Setting & Cases & Subset prediction & Label prediction\\
\midrule\endhead
Planted & 1564 & 0.93 [0.92, 0.95] & 0.89\\
Unplanted & 485 & 0.94 [0.92, 0.96] & 0.89\\
Among Us & 130 & 0.81 [0.74, 0.87] & 0.64\\
\bottomrule\end{longtable}
\endgroup

Second-reader transfer tests whether a selection remains effective on the original actor. On 607 jointly eligible harmful held-out boards, the Qwen reader's intervention leaves 0.10 [0.08, 0.12] violations on Llama, compared with 0.09 after Llama's own selection. Their interventions coincide on 0.66 and certified groups on 0.61. Qwen uses the same support thresholds and its own harmless-twin null calibration. Transfer therefore concerns tested selections under distinct reader scores.

Upstream replay validates one level of relay intervention in the planted setting. Teacher-forced likelihood of the observed relay text supplies the upstream score, evaluated with and without its four available sources. The planted pair lies within the upstream group on 0.97 [0.96, 0.98] of boards, while exact group recovery reaches 0.20 [0.17, 0.22]. Tracing therefore often includes additional sources that support other parts of the reply. Deleting the traced intervention and regenerating the relay leaves 0.05 downstream violations on eligible harmful relayed boards, compared with 0.07 after direct downstream intervention. Four upstream sources fall below the six-unit necessity-screen threshold.

Uncertainty intervals treat the decision's source cluster as the experimental unit. Planted results use 2,000 board bootstrap resamples, except the detection tables, which resample 1,000 harmful and decoy pairs, replay in operational worlds uses 2,000 episode resamples, game results use 2,000 game resamples, and Terrarium uses 2,000 run resamples. Alarm analyses use 1,000 episode resamples with threshold recalibration. Multiple continuations from the same decision are dependent observations within these clusters.

\end{document}